\documentclass{article}

\usepackage{microtype}
\usepackage{graphicx}
\usepackage{subfigure}
\usepackage{booktabs} % for professional tables
\usepackage{url}
\usepackage{makecell}
\usepackage{algorithm}  
\usepackage{algorithmic}  
\usepackage{color}
\usepackage{amsfonts}
\usepackage{amsmath}
\usepackage{eqparbox}
\usepackage{hyperref}

\usepackage[accepted]{icml2021}

\icmltitlerunning{MetaCURE: Meta Reinforcement Learning with Empowerment-Driven Exploration}

\begin{document}

\twocolumn[
\icmltitle{MetaCURE: Meta Reinforcement Learning with \\ Empowerment-Driven Exploration}

% It is OKAY to include author information, even for blind
% submissions: the style file will automatically remove it for you
% unless you've provided the [accepted] option to the icml2021
% package.

% List of affiliations: The first argument should be a (short)
% identifier you will use later to specify author affiliations
% Academic affiliations should list Department, University, City, Region, Country
% Industry affiliations should list Company, City, Region, Country

% You can specify symbols, otherwise they are numbered in order.
% Ideally, you should not use this facility. Affiliations will be numbered
% in order of appearance and this is the preferred way.
\icmlsetsymbol{equal}{*}

\begin{icmlauthorlist}

\icmlauthor{Jin Zhang}{equal,IIIS}
\icmlauthor{Jianhao Wang}{equal,IIIS}
\icmlauthor{Hao Hu}{IIIS}
\icmlauthor{Tong Chen}{IIIS}
\icmlauthor{Yingfeng Chen}{NetEase}
\icmlauthor{Changjie Fan}{NetEase}
\icmlauthor{Chongjie Zhang}{IIIS}
\end{icmlauthorlist}

\icmlaffiliation{IIIS}{Institute for Interdisciplinary Information Sciences, Tsinghua University, China}
\icmlaffiliation{NetEase}{Fuxi AI Lab, NetEase, China}

\icmlcorrespondingauthor{Jin Zhang}{jin-zhan20@mails.tsinghua.edu.cn}

% You may provide any keywords that you
% find helpful for describing your paper; these are used to populate
% the "keywords" metadata in the PDF but will not be shown in the document
\icmlkeywords{Machine Learning, ICML, Meta Learning, Reinforcement Learning, Exploration}

\vskip 0.3in
]

% this must go after the closing bracket ] following \twocolumn[ ...

% This command actually creates the footnote in the first column
% listing the affiliations and the copyright notice.
% The command takes one argument, which is text to display at the start of the footnote.
% The \icmlEqualContribution command is standard text for equal contribution.
% Remove it (just {}) if you do not need this facility.

%\printAffiliationsAndNotice{}  % leave blank if no need to mention equal contribution
\printAffiliationsAndNotice{\icmlEqualContribution} % otherwise use the standard text.

\begin{abstract}
		Meta reinforcement learning (meta-RL) extracts knowledge from previous tasks and achieves fast adaptation to new tasks. Despite recent progress, efficient exploration in meta-RL remains a key challenge in sparse-reward tasks, as it requires quickly finding informative task-relevant experiences in both meta-training and adaptation. To address this challenge, we explicitly model an exploration policy learning problem for meta-RL, which is separated from exploitation policy learning, and introduce a novel empowerment-driven exploration objective, which aims to maximize information gain for task identification. We derive a corresponding intrinsic reward and develop a new off-policy meta-RL framework, which efficiently learns separate context-aware exploration and exploitation policies by sharing the knowledge of task inference. Experimental evaluation shows that our meta-RL method significantly outperforms state-of-the-art baselines on various sparse-reward MuJoCo locomotion tasks and more complex sparse-reward Meta-World tasks.
\end{abstract}

\section{Introduction}
	\label{intro}
	Human intelligence is able to transfer knowledge across tasks and acquire new skills within limited experiences. Current reinforcement learning (RL) agents often require a far more amount of experiences to achieve human-level performance \citep{hessel2018rainbow,vinyals2019grandmaster,hafner2019dream}. To enable sample-efficient learning, meta reinforcement learning (meta-RL) methods have been proposed, which automatically extract prior knowledge from previous tasks and achieve fast adaptation in new tasks \citep{schmidhuber1995learning,finn2017model}. 
	%However, in meta-RL, efficient exploration is vital for both meta-training and adaptation \citep{gupta2018meta}, and becomes a key challenge, especially in sparse-reward tasks.
	Despite fast progress, meta-RL with sparse rewards remains challenging, as task-relevant information is scarce in such settings, and efficient exploration is required to quickly find the most informative experiences in both meta-training and fast adaptation.
The problem of learning effective exploration strategies has been extensively studied for meta-RL with dense rewards,
%\citep{duan2016rl,finn2017model,Stadie2018SomeCO,rothfuss2019promp,zintgraf2019varibad}.
such as E-MAML \citep{Stadie2018SomeCO}, ProMP \citep{rothfuss2019promp} and VariBAD \citep{zintgraf2019varibad}. 
%In these methods, exploration policy learning and exploitation policy learning share the same objective, which is not effective for general sparse-reward tasks. %\citep{duan2016rl,finn2017model,Stadie2018SomeCO,rothfuss2019promp,zintgraf2019varibad}.
Recently, several works have investigated learning to explore in sparse-reward tasks. 
For example, MAESN \citep{gupta2018meta} learns temporally-extended exploration behaviors by injecting structured noises into the exploration policy. PEARL \citep{rakelly2019efficient} explores by posterior sampling \citep{thompson1933likelihood,osband2013more}, which is optimal in asymptotic performance \citep{leike2016thompson}. However, both methods assume access to dense rewards during meta-training.

To address the challenge of meta-RL with sparse rewards, in this paper, we explicitly model the problem of learning to explore, and separate it from exploitation policy learning. This separation allows the learned exploration policy to purely focus on collecting the most informative experiences for enabling efficient meta-training and adaptation. As the common task-inference component in context-based meta-RL algorithms
%(such as PEARL \citep{rakelly2019efficient} and VariBAD \citep{zintgraf2019varibad})
extracts task information from experiences \citep{rakelly2019efficient,zintgraf2019varibad,humplik2019meta}, the exploration policy should collect experiences that contain rich task-relevant information and support efficient task inference. By leveraging this insight, we design a novel empowerment-driven exploration objective that aims to maximize the mutual information between exploration experiences and task identification. We then derive an insightful intrinsic reward from this objective, which is related to model prediction.

To incorporate our efficient exploration method, we develop a new off-policy meta-RL framework, called Meta-RL with effiCient Uncertainty Reduction Exploration (MetaCURE). MetaCURE performs probabilistic task inference, and learns separate context-aware exploration and exploitation policies.
As both context-aware policies depend on task information extracted from the context, MetaCURE shares a common component of task inference for their learning, greatly improving learning efficiency. 
In the adaptation phase, the exploration policy is intrinsically motivated to perform sequential exploratory behaviors across episodes, while the exploitation policy maximizes expected extrinsic return in the last episode of adaptation phase.
	%Moreover, in meta-training, training data is collected by iteratively performing task inference and executing the two policies.
	%MetaCURE enables meta-training with off-policy trajectories, as the gap between off-policy meta-training and on-policy adaptation is alleviated by learning a task inference component that performs inference from on-policy data\citep{rakelly2019efficient}.
	%As MetaCURE separates task inference and policy execution \citep{rakelly2019efficient}, the exploration and exploitation policy can be off-policy trained. This enables collected trajectories to be shared for meta-training, enabling the policies to learn from each other's experiences and achieves superior sample efficiency. 
MetaCURE is extensively evaluated on various sparse-reward MuJoCo locomotion tasks as well as sparse-reward Meta-World tasks. Empirical results show that it outperforms baseline algorithms by a large margin. To illustrate the advantages of our algorithm, we also visualize how it explores during adaptation and compare it with baseline algorithms.

\section{Background}
The field of meta-RL deals with a distribution of tasks $p({\kappa})$, with each task ${\kappa}$ modelled as a Markov Decision Process (MDP), which consists of a state space, an action space, a transition function and a reward function \citep{sutton2018reinforcement}. In common meta-RL settings \citep{duan2016rl,finn2017model,zintgraf2019varibad}, tasks differ in the transition and/or reward function, so we can describe a task ${\kappa}$ with a tuple $\langle p_0^{{\kappa}}(s_0), p^{{\kappa}}(s'|s,a), r^{{\kappa}}(s,a) \rangle$, whose components denote the initial state distribution, the transition probability  and the reward function, respectively. Off-policy meta-RL \citep{rakelly2019efficient} assumes access to a batch of meta-training tasks $\{{\kappa}_m\}_{m=1,2,...,M}$, with $M$ the total number of meta-training tasks. With a slight abuse of notations, we further denote ${\kappa}$ as the task identification, and $\mathcal{K}$ indicates the random variable representing $\kappa$.

We denote context $c_n=(s_n,a_{n},r_{n},s_{n+1})$ as an experience collected at timestep $n$, and $c_{:t}=\left \langle c_{0}, c_{1},..., c_{t-1}\right \rangle$\footnote{For the clarity of following derivations, we define $c_{:0}=\left \langle c_{-1}\right \rangle=\left \langle (\Vec{0},\Vec{0},\Vec{0},s_{0})\right \rangle$.} indicates all experiences collected during $t$ timesteps. Note that $t$ may be larger than the episode length, and when it is the case, $c_{:t}$ represents experiences collected across episodes. $C_{:t}$ denotes a random variable representing $c_{:t}$.
%Similarly, we denote $a_{:t}=\left \langle a_0,a_1,...,a_{t-1}\right \rangle$ as the action sequence in the $t$ timesteps.
%$S^i_{0}$ indicates the random variable representing the $i$-th episode's initial state $s^{i}_{0}$. 

A common objective for meta-RL is to optimize final performance after few-shot adaptation \citep{finn2017model,gupta2018meta,Stadie2018SomeCO,rothfuss2019promp,gurumurthy2020mame}.
%, which is similar to the best arm identification objective in muti-arm bandit problems \citep{audibert2010best}. 
During adaptation, an agent first utilizes some exploration policy $\pi_e$ to explore for a few episodes, and then updates an exploitation policy $\pi$ to maximize the expected return.
% Assume that the agent is given $N$ episodes to perform adaptation, it explores in the first $(N-1)$ episodes, and maximizes expected return in the last episode by extracting and utilizing information from exploration trajectories. 
Such a meta-RL objective can be formulated as:
\begin{equation}
    \max\limits_{\pi,\pi_e} \mathbb{E}_{{{\kappa}}\sim p(\mathcal{K})}[R({{\kappa}},\pi(c^{{\kappa}}_{\pi_e}))],
\end{equation}
where $c^{{\kappa}}_{\pi_e}$ is a set of experiences collected in task $\kappa$ by policy $\pi_e$, and $R({{\kappa}},\pi(c^{{\kappa}}_{\pi_e}))$ is the last episode's expected return with policy $\pi$ in task ${\kappa}$. The policy $\pi$ is adapted with $c^{{\kappa}}_{\pi_e}$ for optimizing final performance. Both $\pi_e$ and $\pi$ can be \textit{context-aware} \citep{lee2020context}, that is, they take context $c_{:i}$ into account while making decisions at timestep $i$.

% As discussed above, current meta-RL algorithms are inefficient in either meta-training exploration or adaptation exploration. Our aim is to improve both kinds of exploration.

%We also care about the sample efficiency of meta-RL algorithms, which can be evaluated by meta-training efficiency (number of samples needed to meta-train the agent) and adaptation efficiency (number of samples needed to acquire good adaptation performance). 

\section{Empowerment-Driven Exploration}
\label{ede}
In this section, we present a novel information-theoretic objective for optimizing exploration in both meta-training and adaptation. In Section \ref{mig}, we derive an insightful intrinsic reward from this exploration objective, which measures experiences' information gain on task identification. 
In Section \ref{help}, we illustrate several implications of this intrinsic reward by a didactic example.

\subsection{Exploration By Maximizing Information Gain}
\label{mig}
To enable efficient meta-RL with sparse rewards, our exploration strategy aims to collect experiences that maximize information gain about the identification of the current task. Consider an exploration policy $\pi_e$ and experiences $C_{:H}$ collected by executing $\pi_e$ for $H$ timesteps in task $\mathcal{K}$. Our exploration objective $\mathcal{J}^{\pi_e}$ is formulated as maximizing the mutual information between exploration experiences $C_{:H}$ and task identification $\mathcal{K}$:
\begin{align}
%\begin{split}
%\begin{aligned}
    &\mathcal{J}^{\pi_e}(C_{:H},\mathcal{K}) \notag \\
    &=I^{\pi_e}(C_{:H};\mathcal{K}) \notag  \\
    %\end{split}
    %\begin{aligned}
        % &= \sum_{i=0}^{N-1}I(C_{iH:(i+1)H-1};\mathcal{T}|C_{0:iH-1},S_0^{i},A_{iH:(i+1)H-1}) \\
        % &= \sum_{i=0}^{N-1}\sum_{j=0}^{H-1}I(C_{iH+j};\mathcal{T}|C_{0:iH+j-1},S_{iH+j},A_{iH+j}) \\
        % &=\sum_{t=0}^{NH-1}I(C_{t};\mathcal{T}|C_{0:t-1},S_{t},A_{t}) \\
        &=\mathbb{E}_{{(c_{:H},\kappa)}\sim (C_{:H},\mathcal{K})}\left[\log\frac{ p^{\pi_e}(c_{:H}|\kappa)}{p^{\pi_e}(c_{:H})}\right] \label{22}\\
        &=\mathbb{E}_{{(c_{:H},\kappa)}\sim (C_{:H},\mathcal{K})} \notag \\ &\ \ \ \ \ \ \ \ \ \ \left[\sum_{t=0}^{H-1}\log \frac{p^{\pi_e}(a_t|c_{:t},\kappa)p(r_{t},s_{t+1}|{\kappa},c_{:t},a_{t})}{p^{\pi_e}(a_t|c_{:t})p(r_{t},s_{t+1}|c_{:t},a_{t})} \right] \label{3}\\
        &=\mathbb{E}_{{(c_{:H},\kappa)}\sim (C_{:H},\mathcal{K})}\left[\sum_{t=0}^{H-1}\log \frac{p(r_{t},s_{t+1}|{\kappa},s_{t},a_{t})}{p(r_{t},s_{t+1}|c_{:t},a_{t})} \right]\notag \\ &\ \ \ \ \ \ \ \ \ \ \ \ \ \ \ \ \ \ \ \ \ \ \ \ \ \ \ \ \ \ \ \ \ \ \ \ \ \ \ \ \ \ \ \ \ \ \ \ \ \ \ \ \ \ \ \ \ \ \ \ \ \ \ \ \ + \textit{Const},
%\end{aligned}
%\end{split}
\label{objective-episode}
\end{align}{}
 where $\textit{Const}$ is a constant that does not change with $\pi_e$. In Eq. \eqref{22}, the mutual information is expressed as the expectation over random variables $C_{:H}$ and $\mathcal{K}$, which follow the probability distribution $p^{\pi_e}(c_{:H},\kappa)=p(\kappa)p^{\pi_e}(c_{:H}|\kappa)$. We then decompose the expectation to each timestep $t$ by the chain rule, as shown in Eq. \eqref{3}. Eq. \eqref{objective-episode} simplifies the expression by taking the following properties: 1) as an exploration policy needs to generalize to new tasks whose identifications are unknown, $\kappa$ is not used for $\pi_e$'s input, and $p^{\pi_e}(a_t|c_{:t},\kappa)=p^{\pi_e}(a_t|c_{:t})$; and 2) according to the Markov property,
%$r_t$ and $s_{t+1}$ are independent from previously collected experiences given $\kappa$, $s_t$ and $a_t$, 
 $p(r_{t},s_{t+1}|{\kappa},c_{:t},a_{t})=p(r_{t},s_{t+1}|{\kappa},s_{t},a_{t})$ for task $\kappa$.
\begin{figure*}[h!]
%\vskip 0.1in
\centering
        \subfigure[Environment illustration.]{\includegraphics[width=0.47\columnwidth]{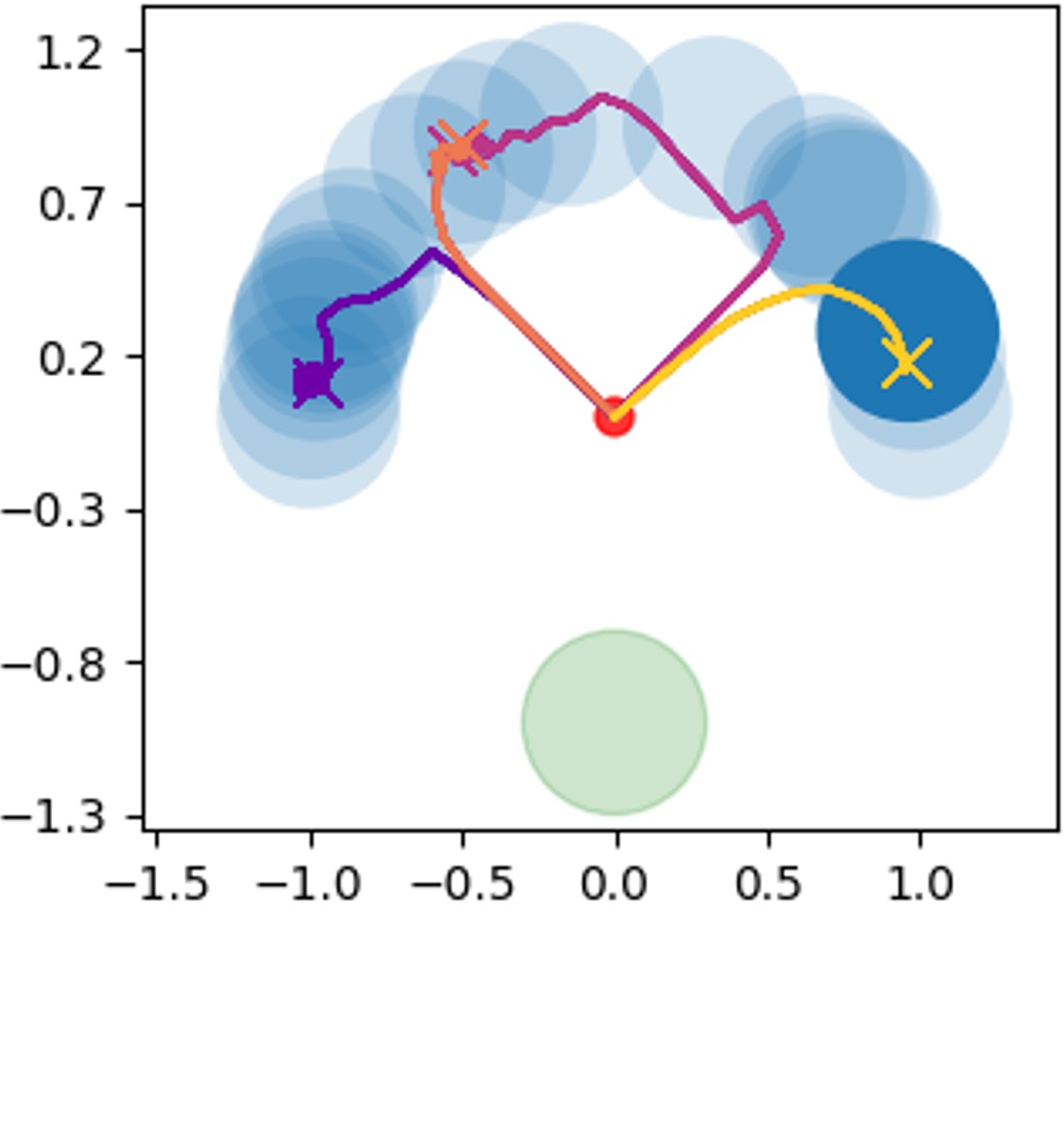}\label{example1}}
        \hspace{0.03\columnwidth}
        \subfigure[Learning curves of $L_{pred}$ and $L_{pred}^{task}$.]{\includegraphics[width=0.45\columnwidth]{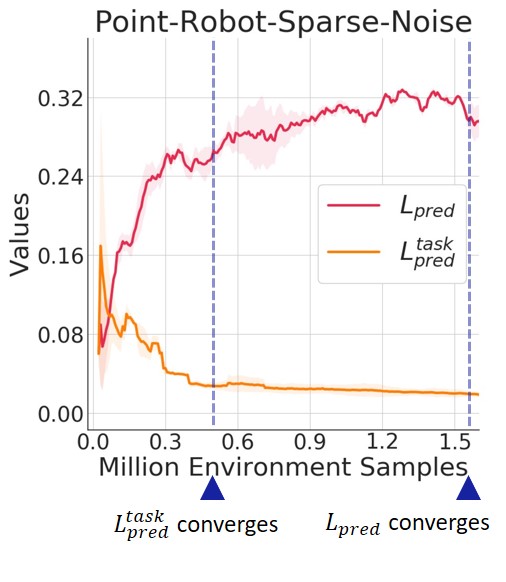}\label{example4}}
        \hspace{0.03\columnwidth}
        \subfigure[Intrinsic rewards at early stages of meta-training.]{\includegraphics[width=0.46\columnwidth]{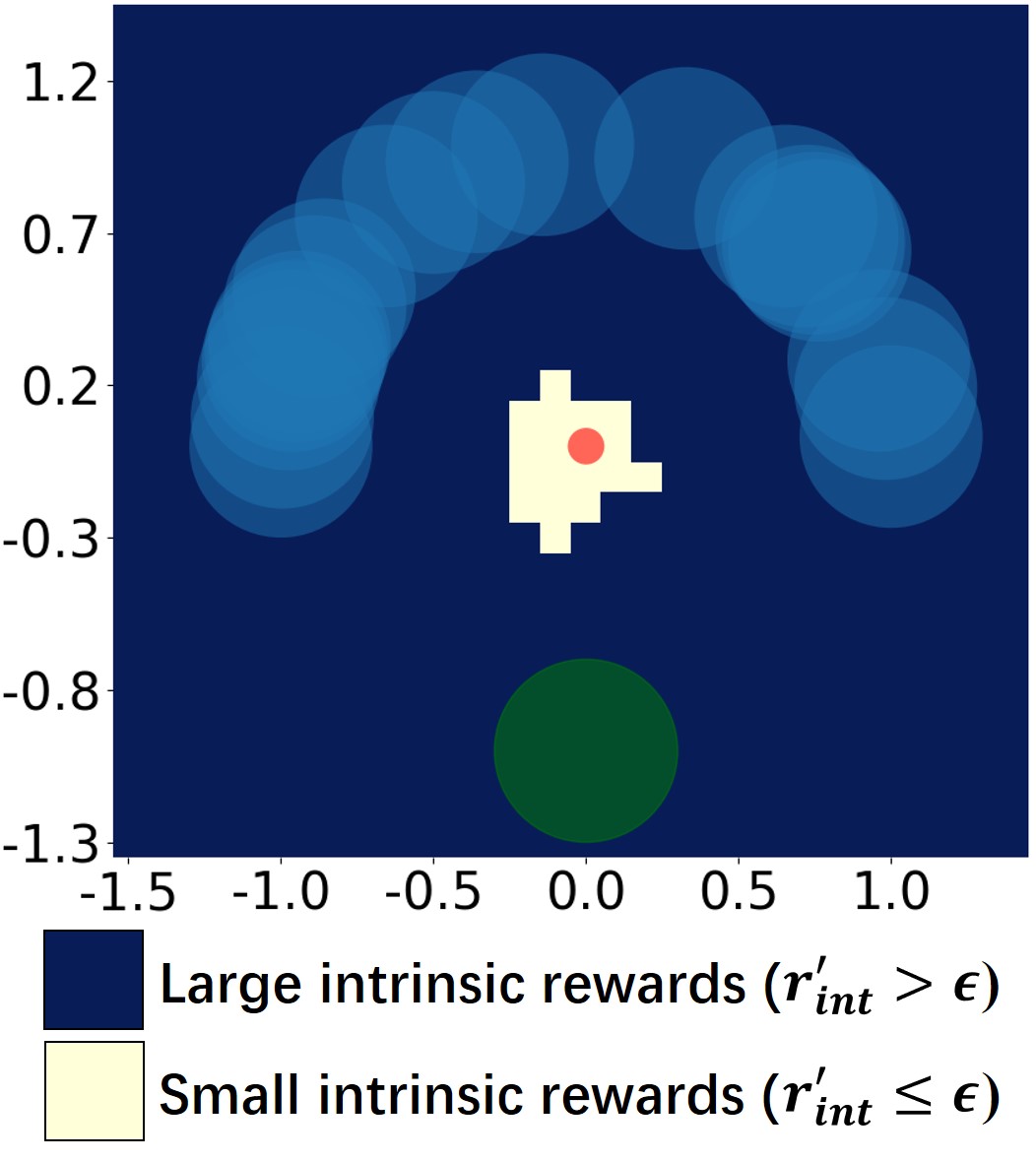}\label{example2}}
        \hspace{0.03\columnwidth}
        \subfigure[Intrinsic rewards at late stages of meta-training.]{\includegraphics[width=0.46\columnwidth]{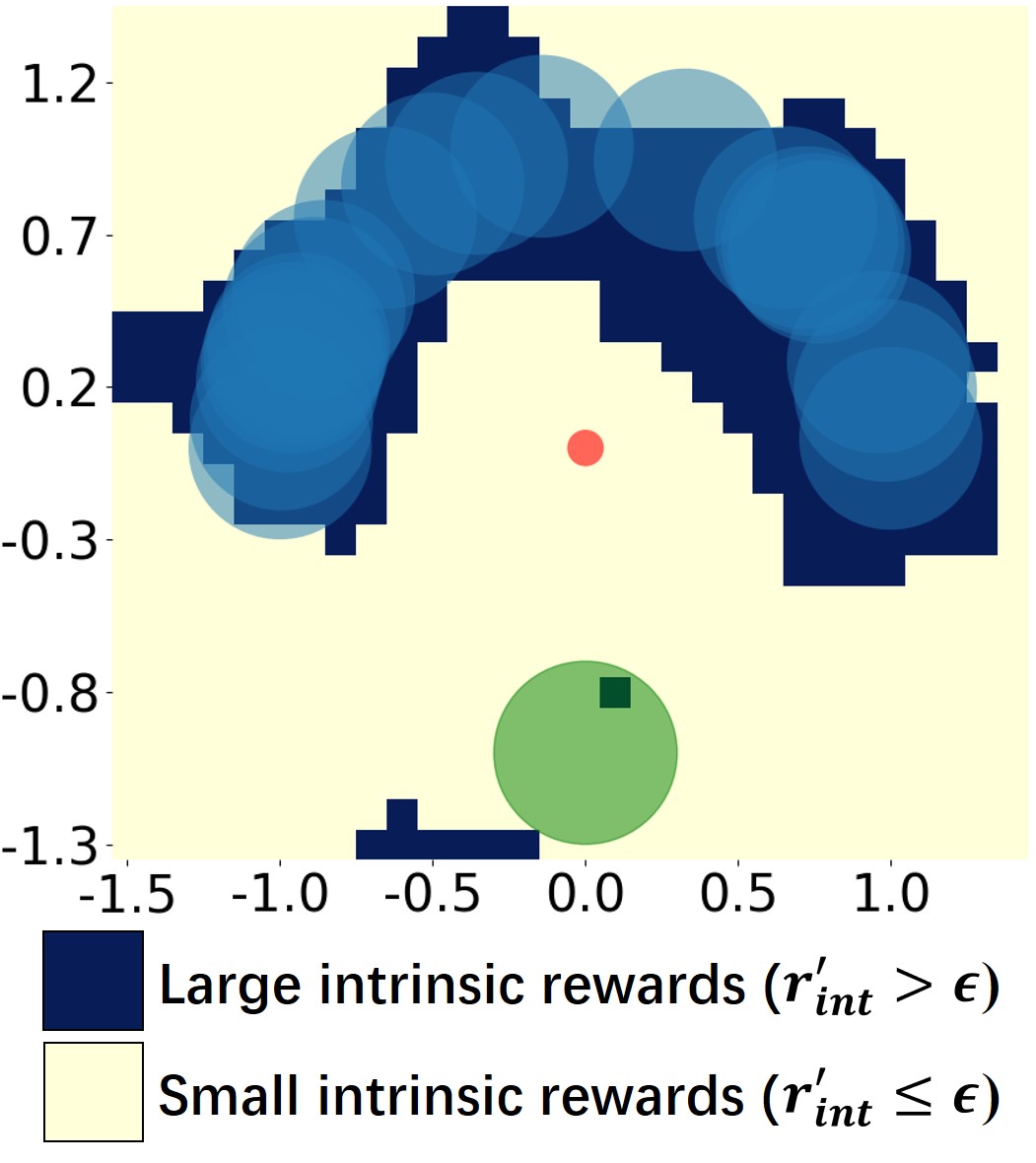}\label{example3}}
\caption{(a) Illustration of the environment. The dark blue circle indicates the goal for the current task, while the light blue circles indicate goals for other meta-testing tasks. The green circle represents noisy regions, and the red circle is agent's initial position. We also illustrate efficient exploration behaviors motivated by our intrinsic reward. The dark purple line is the first episode's trajectory, while the light yellow line represents the last episode's trajectory. (b) Learning curves of the two model prediction errors. $L_{pred}^{task}$ converges much faster (at 0.5 million samples) than $L_{pred}$ (at 1.5 million samples). (c) At early stages of meta-training, our intrinsic reward approximates agent's curiosity about the environment. $\epsilon \textgreater 0$ is a constant to filter out neural network approximation error. Regions around the origin are fully explored, and the intrinsic reward is close to zero. (d) At late stages of meta-training, our intrinsic reward encourages exploration to maximize task information gain. }
\label{example}
%\vskip -0.1in
\end{figure*}
%where $c_{:H}$ and $\kappa$ are jointly sampled from the probability distribution $p_{\pi_e}(c_{:H},\kappa)=p(\kappa)p_{\pi_e}(c_{:H}|\kappa)$. The last equation takes the following properties: 1) as the policy needs to generalize to new tasks whose identifications are unknown, $\kappa$ is not used for $\pi_e$'s input, and $p_{\pi_e}(a_t|c_{:t},\kappa)=p_{\pi_e}(a_t|c_{:t})$. 2) according to the Markov property \citep{sutton2018reinforcement}, $r_t$ and $s_{t+1}$ are independent from previously collected experiences given $\kappa$, $s_t$ and $a_t$, so $p(r_{t},s_{t+1}|{\kappa},c_{:t},a_{t})=p(r_{t},s_{t+1}|{\kappa},s_{t},a_{t})$.

On the right-hand side of Eq. \eqref{objective-episode}, the numerator $p(r_{t},s_{t+1}|{\kappa},s_{t},a_{t})$ indicates the predictability of rewards and transitions given the task identification $\kappa$, while the denominator $p(r_{t},s_{t+1}|c_{:t},a_{t})$ indicates predictability given current context $c_{:t}$. The logarithm of these two terms' division measures the amount of task information that the task identification contains more than current context. Note that the expected log probability can be interpreted as negative cross-entropy prediction loss, so our exploration intrinsic reward is defined as:
% \begin{equation}
% \begin{split}
%         &r_{\text{TURE}}'(c_{0:t},{T_m})\\
%         &=\log p(r_{t},s_{t+1}|{T_m},s_{t},a_{t}) -\log p(r_{t},s_{t+1}|c_{0:t-1},s_{t},a_{t}) \\
%          &=L_{pred}(c_{0:t}) -L_{pred}^{task}({T}_m,c_t),
% \end{split}
% \label{objective}
% \end{equation}{}
\begin{equation}
\begin{split}
        &r_{\text{int}}'(c_{:t+1},{\kappa})\\
        &\ \ \ \ \ \ =\underbrace{-\log p(r_{t},s_{t+1}|c_{:t},a_{t})}_{L_{pred}(c_{:t+1})} +\underbrace{\log p(r_{t},s_{t+1}|{\kappa},s_{t},a_{t})}_{-L_{pred}^{task}(\kappa,c_t)}.
\end{split}
\label{objective}
\end{equation}{}
This intrinsic reward can be interpreted as the difference of two model prediction errors $L_{pred}$ and $L_{pred}^{task}$, which can be estimated by training two model predictors, respectively: the \textit{Meta-Predictor} makes predictions based on the current context, while the \textit{Task-Predictor} makes predictions based on the task identification. To estimate the log probabilities tractably, we follow the common approach of utilizing L2 distances as an approximation of the negative log probability \citep{chung2015recurrent,babaeizadeh2017stochastic}:
\begin{equation}
\begin{split}
  &L_{pred}(c_{:t+1})\approx \left(r_t-\hat{r}_t^{pred}\left(c_{:t},a_t\right)\right)^2\\
  &\ \ \ \ \ \ \ \ \ \ \ \ \ \ \ \ \ \ \ \ \ \ \ \ \ \ \ \ \ \ \ \ \ \ \ \ \  +\left\|s_{t+1}-\hat{s}_{t+1}^{pred}(c_{:t},a_t)\right\|_2^2 \\
  &L_{pred}^{task}(\kappa,c_t)\approx  \left(r_t-\tilde{r}_t^{pred}\left(\kappa,s_t,a_t\right)\right)^2\\
  & \ \ \ \ \ \ \ \ \ \ \ \ \ \ \ \ \ \ \ \ \ \ \ \ \ \ \ \ \ \ \ \ \ \ \ \ \  +\left\|s_{t+1}-\tilde{s}_{t+1}^{pred}(\kappa,s_t,a_t)\right\|_2^2,
\end{split}
\label{intrinsicr2}
\end{equation}{}
where $\hat{r}_t^{pred}$ and $\hat{s}_{t+1}^{pred}$ are reward and transition predicted by the Meta-Predictor, while $\tilde{r}_t^{pred}$ and $\tilde{s}_{t+1}^{pred}$ are predicted by the Task-Predictor. Proofs of this section are deferred to Appendix A.
% \footnote{}.

\subsection{Didactic Example}
\label{help}
In this section, we demonstrate the underlying implications of our intrinsic reward $r_{\text{int}}'$ with detailed analyses on a didactic example.

We propose a 2-D navigation task set called Point-Robot-Sparse-Noise, as shown in Figure \ref{example1}. The agent's observation is a 3-D vector composed of its current position $(x,y)$ and a noise term $u$. $u$ is a Gaussian noise if the agent is located in the green circle, and is zero otherwise. Goals are uniformly distributed on a semicircle of radius 1 and only sparse reward is provided. To understand how our intrinsic reward works, we illustrate its values during the learning process, as shown in Figure \ref{example}. This illustration demonstrates the following properties of our intrinsic reward:
% , i.e., $r_{intr}'(c_0,s,a,r,s',\mathcal{T}_m)$

% \textbf{Bonus on meta-training exploration} Although our intrinsic reward $r_{int}'$ does not directly optimize for meta-training exploration, in early stages of training, it approximately reflects the agent's curiosity \citep{schmidhuber1991possibility} about the environment and encourages meta-training exploration. This is because that the baseline term $L_{pred}^{oracle}$ converges much faster than $L_{pred}$, as it is aware of the true task identification and obtains a simple input space. Thus, $r_{int}'$ approximately equals to  $L_{pred}$, which can be viewed as a curiosity-driven intrinsic reward \citep{burda2018large,burda2018exploration}. As shown in Figure \ref{example2}, in early stages of training, $r_{int}'$ is positive in unexplored regions, encouraging meta-training exploration.

\begin{figure*}[ht]
%\vskip 0.2in
\centering
        %\hspace{0.09\columnwidth}
        \subfigure{\includegraphics[width=1.3\columnwidth]{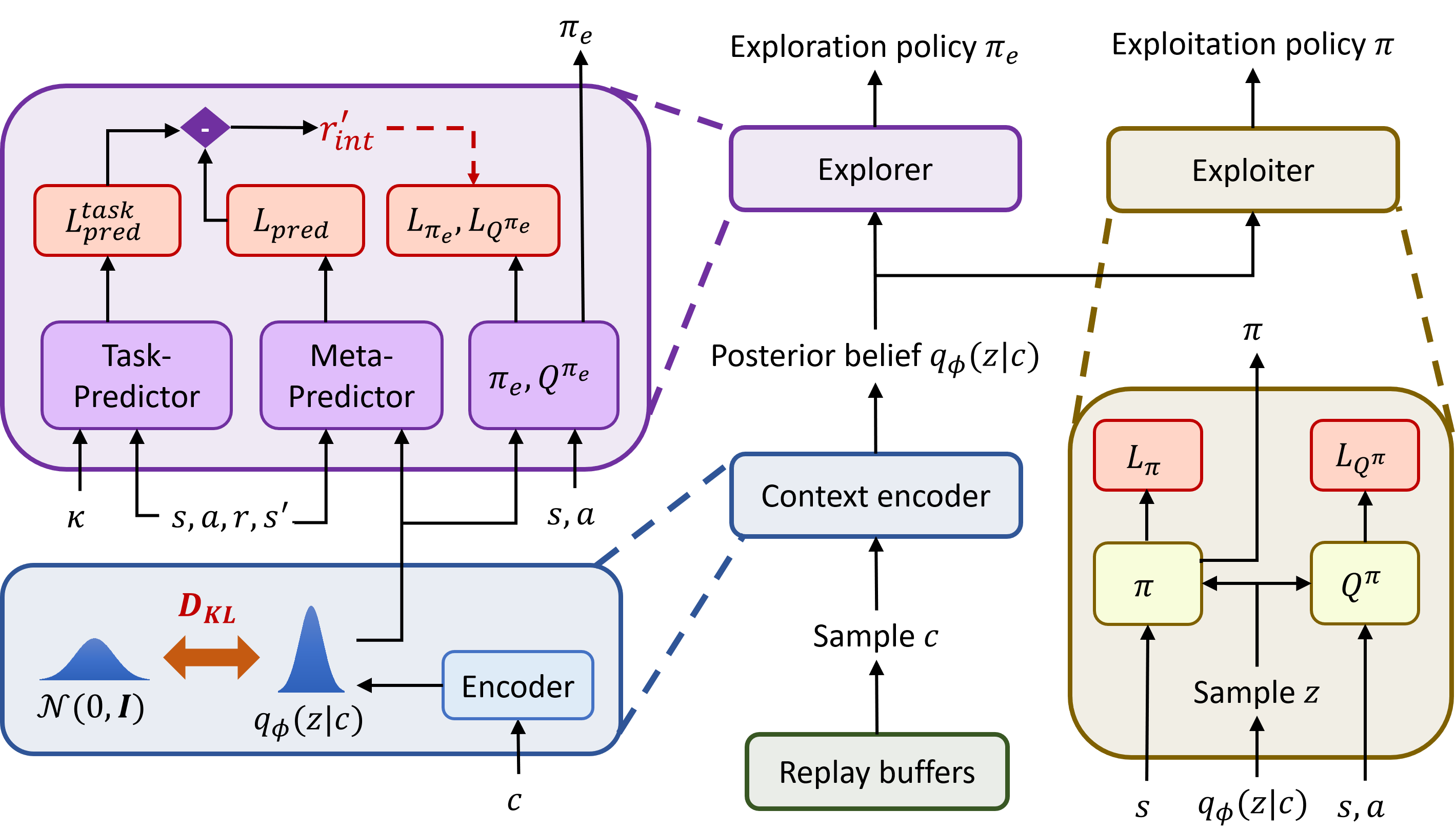}}
\caption{MetaCURE's meta-training pipeline. $L_{\pi_e}$, $L_{Q^{\pi{_e}}}$, $L_{\pi}$ and $L_{Q^{\pi{}}}$ are corresponding SAC loss functions for the exploration and exploitation policies.}
\label{algorithm}
%\vskip -0.2in
\end{figure*}

\textbf{Approximate curiosity-driven exploration at early stages of meta-training:} in order to learn an efficient exploration policy, a meta-RL agent needs to effectively explore the tasks during meta-training. We found that at early stages of meta-training, our intrinsic reward demonstrates approximate curiosity-driven exploration \citep{schmidhuber1991possibility}. This is because that the second model prediction error $L_{pred}^{task}$ converges much faster than $L_{pred}$, as shown in Figure \ref{example4}, as it utilizes extra information by being aware of the task identity. Thus, $r_{\text{int}}'$ approximately equals to $L_{pred}$, which can be viewed as a curiosity-driven intrinsic reward \citep{burda2018large,burda2018exploration} measured by model prediction errors \citep{pathak2017curiosity}. As shown in Figure \ref{example2}, at early stages of meta-training, $r_{\text{int}}'$ obtains large values in unexplored regions, encouraging visitation to these regions.

% \textbf{Task uncertainty reduction exploration during adaptation} After the estimation of $r_{int}'$ converges, it encourages task inference during adaptation (Figure \ref{example3}), assigning positive intrinsic rewards to potentially rewarding regions that help identify tasks. This is exactly the objective from which the intrinsic reward is derived (Equation \ref{objective-episode}): collecting experiences that are maximally informative for inferring current task.

% \textbf{Task uncertainty reduction exploration during adaptation:} after the estimation of $r_{int}'$ converges, positive intrinsic rewards are assigned to potentially rewarding regions that help identify tasks, encouraging exploration behaviors that reduce task uncertainty during adaptation, as shown in Figure \ref{example3}. This is exactly the objective from which the intrinsic reward is derived (Equation \ref{objective-episode}): collecting experiences that are maximally informative for inferring current task.

\textbf{Empowerment-driven exploration at late stages of meta-training:} in both meta-training and adaptation, the exploration policy needs to collect experiences that contain rich task information. At late stages of training, the intrinsic reward encourages collection of experiences that are informative for inferring the current task, as it is derived from the objective of maximizing information gain about task identification (Eq. \eqref{objective-episode}). As shown in Figure \ref{example3}, $r_{\text{int}}'$ is large in regions where goals are possibly located, which shows that our intrinsic reward has implicitly learned the task distribution and encourages efficient exploration to acquire task information.

\renewcommand\algorithmiccomment[1]{%
  \hfill$\triangleright$\ \eqparbox{COMMENT}{#1}%
}

\textbf{Robustness to irrelevant noises:} Tasks may obtain uninformative noises that distract the agent from efficient exploration. At late stages of meta-training, $r_{\text{int}}'$ not only encourages the agent to focus on uncertainties that help inference of task identification, but also ignores irrelevant noises. This is achieved by subtracting the second term $L_{pred}^{task}$. $L_{pred}^{task}$ measures uncertainty given the true task identification, which is not helpful for inferring current task, e.g., the well-known noisy TV problem \citep{burda2018exploration}. In our example, at late stages of meta-training (Figure \ref{example3}), in the noisy green circle, the mean of $L_{pred}$ is 1.306, and the mean of $L_{pred}^{task}$ is 1.312. In contrast, the mean of $r_{\text{int}}'$ is only -0.006. Thus, the agent no longer explores the noisy regions.

\section{MetaCURE Framework}
\label{method}
This section presents MetaCURE, a new off-policy context-based meta-RL framework that learns separate exploration and exploitation policies.
%Our exploration objective proposed in Section \ref{ede} is different from the common exploitation objective of maximizing expected extrinsic returns \citep{duan2016rl,zintgraf2019varibad}, so we learn separate exploration and exploitation policies. 
It also performs probabilistic task inference, which captures uncertainty over the task.
%We utilize variational inference methods to infer posterior task belief. The exploration policy is motivated by our intrinsic reward to collect informative experiences in adaptation, and the exploitation policy maximizes the expected extrinsic return in the last episode of adaptation. The exploration and exploitation policies share a common replay buffer, enabling them to learn from each other's experiences. We also use a separate buffer for sampling contexts.
As shown in Figure \ref{algorithm}, MetaCURE is composed of four main components: (i) a context encoder $q_{\phi}(z|c)$ that extracts task information from context $c$ and infers the posterior belief over the task embedding $z$, (ii) an \textit{Explorer} that learns a context-aware exploration policy $\pi_e$, as well as two model predictors used to estimate our intrinsic reward, (iii) an \textit{Exploiter} that learns a context-aware exploitation policy $\pi$, and (iv) replay buffers storing training data. Our exploration objective proposed in Section \ref{ede} is different from the common exploitation objective of maximizing expected extrinsic returns \citep{duan2016rl,zintgraf2019varibad}, so we learn separate exploration and exploitation policies. As for the context encoder and the Exploiter, we adopt a variational inference structure similar to PEARL \citep{rakelly2019efficient}. During meta-training, data is collected by iteratively inferring the posterior task belief with contexts and performing both exploration and exploitation policies. During adaptation, only the exploration policy is used to collect informative experiences for task inference, and the exploitation policy utilizes this posterior belief to maximize final performance.

\begin{algorithm}[t]
   \caption{MetaCURE: Meta-training Phase}
   \label{alg:MetaCURE}
\begin{algorithmic}
   \STATE {\bfseries Require:} A set of meta-training tasks $\{{\kappa}_m\}_{m=1,2,...,M}$ drawn from $p(\mathcal{K})$
   \STATE Initialize replay buffers $\mathcal{B}^{{\kappa}_m}$ for each task ${\kappa}_m$ %\Comment{Test}
   \STATE Initialize exploration policy $\pi_e$, exploitation policy $\pi$, context encoder $q_{\phi}$, Task-Predictor and Meta-Predictor
   \WHILE{not done} 
   \FOR[Data collection]{each task ${{\kappa}_m}$}
%   \STATE Initialize context $c^{{\kappa}_m}=\{\}$
%   \STATE Collect $E_1$ episodes' experiences by iteratively rolling out $\pi_e$ and updating $q_{\phi}(z|c^{{\kappa}_m})$
%   \STATE Add collected experiences to $\mathcal{B}^{{\kappa}_m}$ and $\mathcal{B}_{enc}^{{\kappa}_m}$
   \STATE Collect exploration and exploitation experiences by running Algorithm \ref{MetaCURE adaptation} on ${\kappa}_m$
   \STATE Add collected experiences to $\mathcal{B}^{{\kappa}_m}$
%   \FOR[Explorer rollout]{episodes=1,...,$E_1$}
%   \FOR{steps=1,...,$T$}
%   \STATE Take action according to $\pi_e(a|s,q_\phi(z|c_m))$
   %\STATE Add collected experiences to $\mathcal{B}^{{\kappa}_m}$
%   \ENDFOR
%   \ENDFOR
%   \STATE Collect $E_2$ episodes' experiences by iteratively rolling out $\pi$ and sampling $z$ from $q_{\phi}(z|c^{{\kappa}_m})$
%   %\FOR[Exploiter rollout]{episodes=1,...,$E_2$}
%   %\STATE Sample $z\sim q_\phi(z|c_m)$
%   \STATE Add collected experiences to $\mathcal{B}^{{\kappa}_m}$
   %\ENDFOR
   \ENDFOR
   \FOR[Training]{steps in training steps}
   \FOR{each ${\kappa}_m$}
   \STATE Sample context batch $b_{enc}^{{\kappa}_m}$ from $\mathcal{B}^{{\kappa}_m}$
   \STATE Sample policy training batch $b^{{\kappa}_m}$ from $\mathcal{B}^{{\kappa}_m}$
   \STATE Train Task-Predictor and Meta-Predictor by minimizing $L_{pred}^{task}$ and $L_{pred}$ in Eq. \eqref{intrinsicr2}, respectively
   \STATE Train context encoder $q_{\phi}$ by maximizing Eq. \eqref{elbo}
   \STATE Compute $\pi_e$'s reward $r_{\text{exploration}}$ using Eq. \eqref{exploration reward}
   %\STATE Train $q_{\phi}$, $\pi$ and $\pi_e$ with corresponding loss functions, train the predictors to minimize prediction loss on $b^m$
   \STATE Train $\pi$ and $\pi_e$ by minimizing corresponding SAC losses in Eq. \eqref{sacloss2}
   \ENDFOR
   \ENDFOR
   \ENDWHILE
\end{algorithmic}
\label{MetaCURE meta-training}
\end{algorithm}
\begin{algorithm}[t]
   \caption{MetaCURE: Adaptation Phase}
\begin{algorithmic}
   \STATE {\bfseries Require:} Meta-test task drawn from $p({\mathcal{K}})$, number of adaptation episodes $E$
   \STATE Initialize context $c=\{\}$
   \FOR[Exploration phase]{episodes=1,...,$E-1$}
   \FOR{steps=1,...,$T$}
   \STATE Take action according to $\pi_e(a|s,q_\phi(z|c))$
   \STATE Add collected experience $(s,a,r,s')$ to $c$
   \ENDFOR
   \ENDFOR
    \STATE Sample $z\sim q_\phi(z|c)$
   \FOR[Exploitation phase]{steps=1,2,...,$T$}
   \STATE Take action according to $\pi(a|s,z)$
   \ENDFOR
\end{algorithmic}
\label{MetaCURE adaptation}
\end{algorithm}

\textbf{The context encoder} $q_{\phi}$ uses variational inference methods \citep{kingma2013auto,alemi2016deep} to model the exploitation policy $\pi$'s state-action value function $Q^{\pi}$. The evidence lower bound for training the context encoder is:
\begin{equation}
\begin{split}
  &\mathbb{E}_{z\sim q_\phi(z|c),s,a}[\log p(Q^{\pi}(s,a,z))\\
  &\ \ \ \ \ \ \ \ \ \ \ \ \ \ \ \ \ \ \ \ \ \ \ \ \ \ \ \ \ \ \ \ \ \ \ \ \ \ \ \ \ \ \ \ \ \ \ \ -\beta D_{KL}(q_{\phi}(z|c)||p(z))] ,
\end{split}
\label{elbo}
\end{equation}{}
where $\beta$ is a hyper-parameter, $s$, $a$ and $c$ are sampled from the replay buffers. In practice, to tractably estimate the log probability of $Q^{\pi}$, we follow PEARL's method of utilizing negative TD errors as an approximation of the log probability.

% $q_{\phi}$ utilizes variational inference methods \citep{kingma2013auto,alemi2016deep} to estimate the posterior belief over task embeddings. In order to learn effective embeddings, its decoder is designed to model the exploitation policy $\pi$'s state-action value function \citep{rakelly2019efficient}, which captures rich and temporally-extended information about the current task.

%Intuitively, this task inference component can be used by both the exploration and exploitation policy, as it only aims at extracting useful task information \citep{rakelly2019efficient,zintgraf2019varibad}. So in MetaCURE, the task encoder can be reused by the exploration policy to infer posterior task belief, even if the encoder does not get gradients from the exploration policy. This knowledge reuse greatly improves computation efficiency.
%$s$, $a$ and $c$ are sampled from the same task.
%The loss function for the task encoder is:
% \begin{align}
% \begin{split}
% \label{encoder loss}
%         L_{enc}=\mathbb{E}_{z\sim q_\phi(z|c)}[\log p()]
%     \end{split}
% \end{align}

\textbf{The Explorer} consists of an exploration policy $\pi_e$ that effectively explores during both meta-training and adaptation, its corresponding state-action value function $Q^{\pi_{e}}$, as well as two model predictors for estimating our intrinsic reward. The reward signal for $\pi_e$ is defined as follows:
\begin{align}
\begin{split}
\label{exploration reward}
        r_{\text{exploration}}(c_{:t+1},\kappa)=r_{\text{int}}'(c_{:t+1},\kappa)+\lambda r_t,
    \end{split}
\end{align}
where $\kappa$ is the task identification, $r_{\text{int}}'$ is our intrinsic reward defined in Eq. \eqref{objective}, $\lambda$\textgreater$0$ is a hyper-parameter, and $r_t$ is the extrinsic reward. $r_{\text{int}}'$ is estimated by training two model predictors and subtracting their prediction errors: the \textit{Task-Predictor} makes predictions based on the task identification $\kappa$, while the \textit{Meta-Predictor} makes predictions based on the context $c$. Both predictors are trained by minimizing their prediction errors in order to estimate $L_{pred}^{task}$ and $L_{pred}$, respectively. Note that the exploration policy learns biased exploration behaviors by incorporating the extrinsic reward, as it effectively supports task inference, and we find empirically that considering extrinsic rewards leads to superior performance.
Both the exploration policy $\pi_e$ and the Meta-Predictor need to extract task information from the context, which can be represented as agent's posterior task belief $q_{\phi}(z|c)$.
Thus MetaCURE takes $q_\phi(z|c)$ instead of context $c$ as the input of both $\pi_e$ and the Meta-Predictor. This knowledge reuse greatly improves learning efficiency.

\textbf{The Exploiter} consists of an exploitation policy $\pi(a|s,z)$ that utilizes exploration experiences to perform exploitation behaviors, as well as its corresponding state-action value function $Q^{\pi}$. We design $\pi$ to take state $s$ and the task embedding $z$ sampled from the posterior task belief $q_{\phi}(z|c)$ as input, and optimizes it with the extrinsic reward $r$. 
% This structure can be explained as context-based meta-RL with probabilistic context embedding \citep{rakelly2019efficient}. 
%Finally, we utilize SAC \citep{haarnoja2018soft}, an off-policy RL algorithm to train both policies. 

\textbf{The replay buffers} $\{\mathcal{B}^{\kappa_{m}}\}_{m=1,2,...,M}$ are used for storing data from the meta-training tasks $\{{\kappa_{m}}\}_{m=1,2,...,M}$.
As off-policy learning enables training on data collected by other policies, and our intrinsic reward can be computed with off-policy experience batches, $\pi_e$ and $\pi$ share the same replay buffers. This allows useful experiences to be shared between policies, greatly improving sample efficiency.

\begin{figure*}[h!]
%\vskip 0.1in
\centering
        \subfigure{\includegraphics[width=0.65\columnwidth]{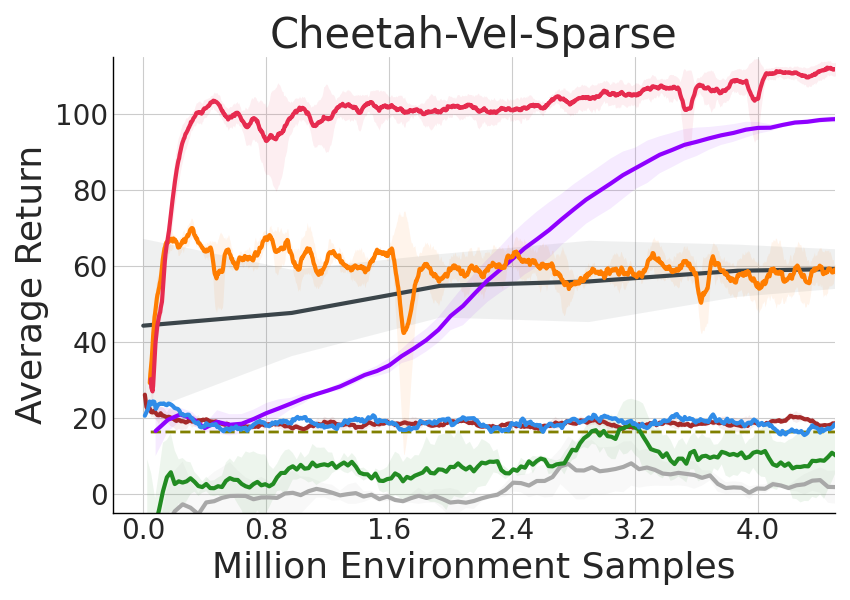}}
        \subfigure{\includegraphics[width=0.65\columnwidth]{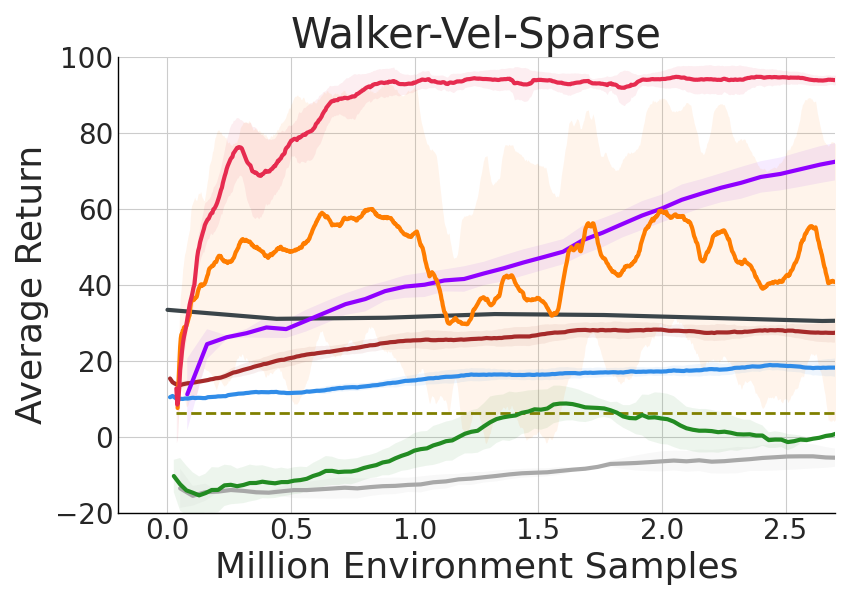}}
        \subfigure{\includegraphics[width=0.65\columnwidth]{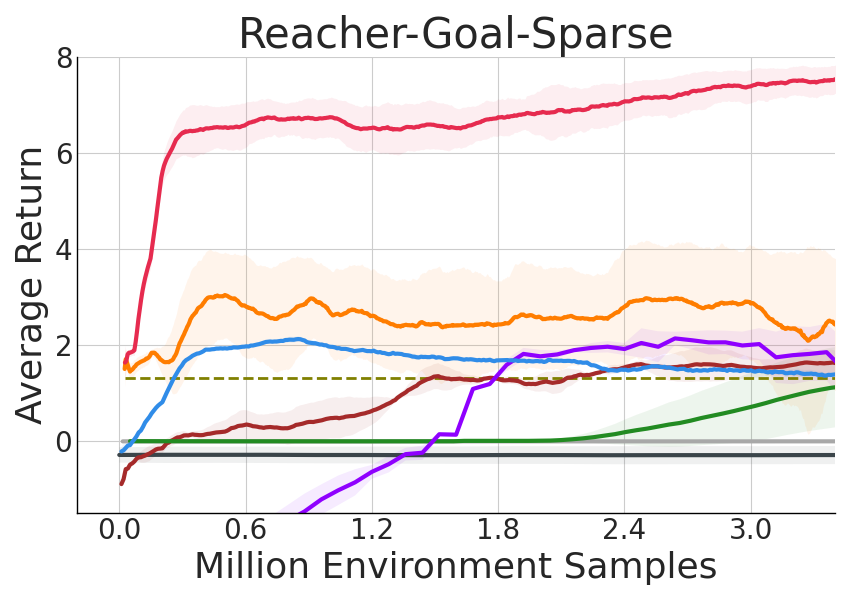}}
        \subfigure{\includegraphics[width=0.65\columnwidth]{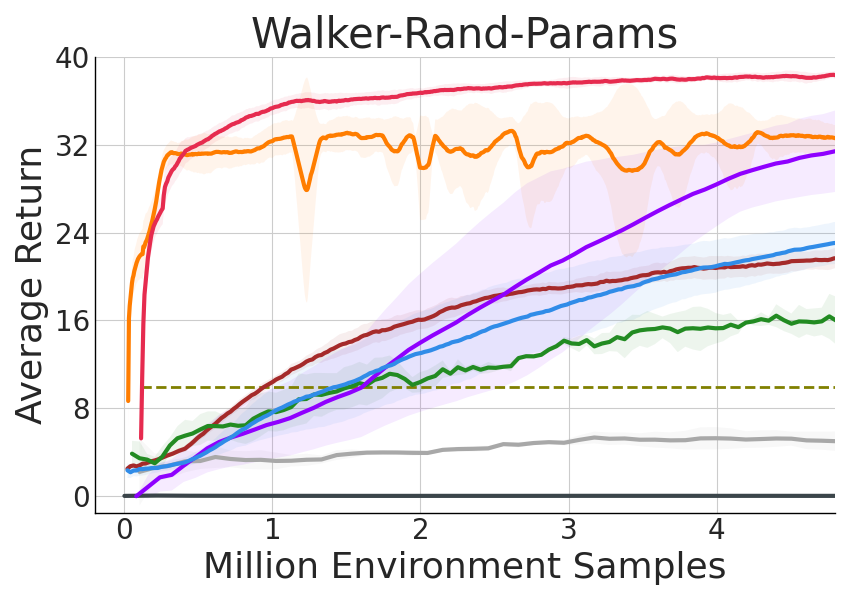}}
        \subfigure{\includegraphics[width=0.65\columnwidth]{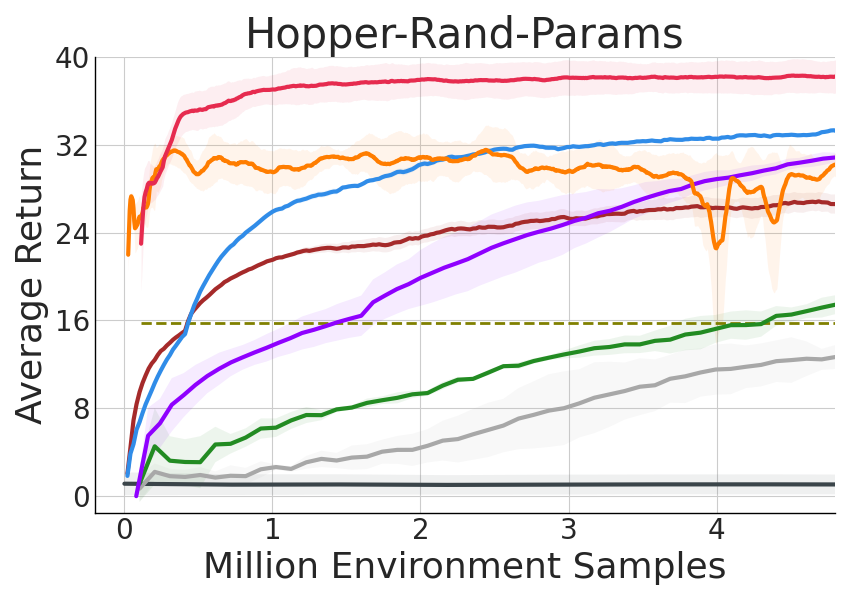}}
        \subfigure{\includegraphics[width=0.65\columnwidth]{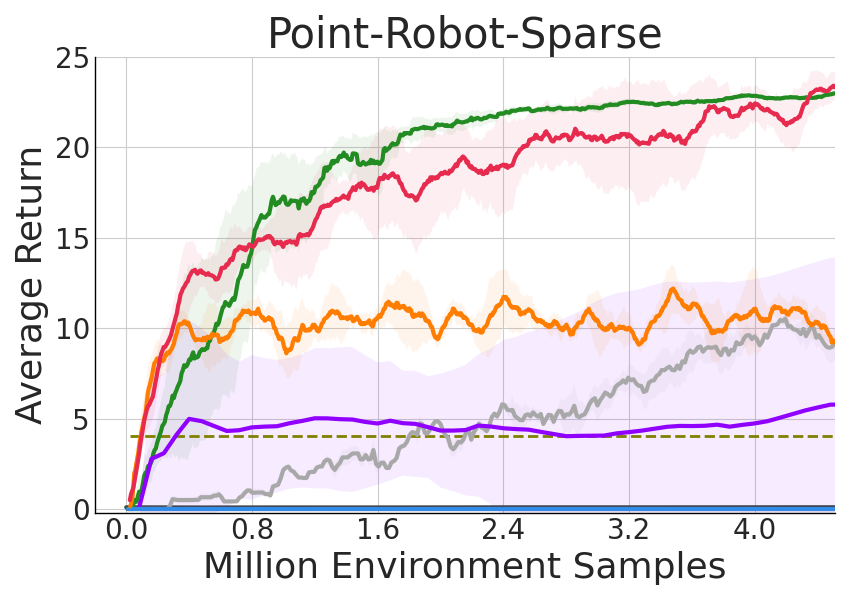}}
        \subfigure{\includegraphics[width=1.35\columnwidth]{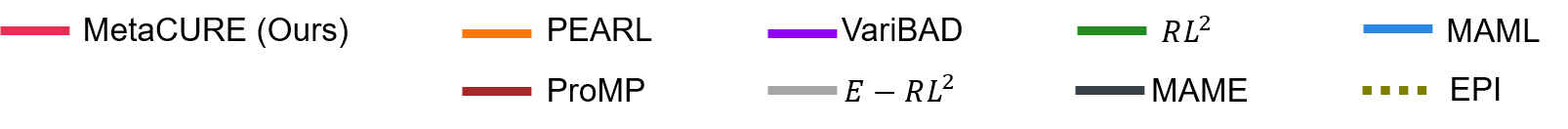}}
\caption{Evaluation of MetaCURE and several meta-RL baselines on various sparse-reward continuous control task sets. We plot the algorithms' meta-testing performance as a function of the number of experiences collected during meta-training. MetaCURE achieves substantially better performance than baseline algorithms.}
\label{curves}
%\vskip -0.1in
\end{figure*}

Both $\pi_e$ and $\pi$ are off-policy trained with SAC \citep{haarnoja2018soft}. As shown in Figure \ref{algorithm}, the loss functions for training the policies are as follows:
\begin{equation}
\begin{split}
  & L_{\pi}=\mathbb{E}_{s,c,z}\left[D_{KL}(\pi(a|s,\overline{z})||\frac{exp(Q^\pi(s,a,\overline{z}))}{\mathcal{Z}_\pi(s,\overline{z})})\right] \\
  &L_{\pi_e}=\mathbb{E}_{s,c}\\
  &\ \ \ \ \ \left[D_{KL}\left(\pi_e(a|s,\overline{q_\phi}(z|c))||\frac{exp(Q^{\pi_e}(s,a,\overline{q_\phi}(z|c)))}{\mathcal{Z}_{\pi_e}(s,\overline{q_\phi}(z|c))}\right)\right]\\
  &L_{Q^{\pi}}=\mathbb{E}_{(s,a,r,s'),c,z}\left[Q^{\pi}(s,a,z)-(r+\gamma\overline{V_{\pi}}(s',\overline{z}))\right]^2 \\
     &L_{Q^{{\pi}_e}}=\mathbb{E}_{(s,a,r,s'),c}\left[ Q^{\pi_e}(s,a,\overline{q_\phi}(z|c)) \right.\\
    &\ \ \ \ \ \ \ \ \ \ \ \ \ \ \ \ \ \ \ \ \ \ \ \ \ \left.-(r_{exploration}+\gamma\overline{V_{\pi_e}}(s',\overline{q_\phi}(z|c')))\right]^2,
\end{split}
\label{sacloss2}
\end{equation}{}
where $\overline{z}$ and $\overline{q_\phi}$ indicate that gradients do not flow through them, $\mathcal{Z}_\pi$ and $\mathcal{Z}_{\pi_e}$ are normalization functions that do not affect gradients, $\overline{V_{\pi}}$ and $\overline{V_{\pi_e}}$ are target value functions, and $c'=c\cup \{(s,a,r,s')\}$ is the updated context. All the expectations over $s$ and $(s,a,r,s')$ are averaged over the replay buffers. $c$ is randomly sampled from exploration experiences collected by $\pi_e$, and all the expectations over $z$ are averaged over the posterior task belief $q_{\phi}(z|c)$. 
Pseudo-codes for meta-training and adaptation are available in Algorithm \ref{MetaCURE meta-training} and Algorithm \ref{MetaCURE adaptation}, respectively. Additional implementation details are deferred to Appendix B. Our implementation codes are available at \url{https://github.com/NagisaZj/MetaCURE-Public}.

\section{Experiments}
In this section, we aim at answering the following questions: 1) Can MetaCURE achieve excellent adaptation performance in sparse-reward tasks that require efficient exploration in both meta-training and adaptation (Section \ref{res1} \& \ref{res2})?
2) Can the exploration policy collect informative experiences efficiently (Section \ref{res3})?
3) Are the intrinsic reward as well as the separation of exploration and exploitation critical for MetaCURE's performance (Section \ref{res4})?
%4. Is the separation of exploration and exploitation vital for good performance?
% Besides, we are also interested in its sample efficiency.

\subsection{Adaptation Performance on Continuous Control}
\label{res1}

\textbf{Environment Setup:} algorithms are evaluated on six continuous control task sets with sparse rewards, in which exploration is vital for superior performance. Tasks vary in either the reward function (goal location in Point-Robot-Sparse and Reacher-Goal-Sparse, target velocity in Cheetah-Vel-Sparse and Walker-Vel-Sparse) or the transition function (Walker-Params-Sparse and Hopper-Rand-Params). These tasks (except for Point-Robot-Sparse) are simulated via MuJoCo \citep{todorov2012Mujoco} and are benchmarks commonly used by current meta-learning algorithms \citep{mishra2017simple,finn2017model,rothfuss2019promp,rakelly2019efficient}. Unlike previous evaluation settings, we limit the length of adaptation phase to investigate the efficiency of exploration. Also, dense reward is not provided in meta-training, which is different from the setting of PEARL.
% Typically, each adaptation phase consists of 2$\sim$4 episodes, each episode 32$\sim$64 steps, varying with the specific task set. 
Detailed parameters and reward function settings are deferred to Appendix C.

% \renewcommand{\arraystretch}{1.3}
% \begin{table*}[ht]
% \caption{Final success rates on the hard sparse-reward Meta-World tasks. Replace with learning curve!}
% \label{metaworld-table}
% %\vskip 0.15in
% \centering
% \begin{center}
% \begin{small}
% \begin{sc}
% \begin{tabular}{p{0.4\columnwidth}|p{0.26\columnwidth}p{0.2\columnwidth}p{0.2\columnwidth}p{0.13\columnwidth}p{0.13\columnwidth}p{0.13\columnwidth}p{0.13\columnwidth}p{0.13\columnwidth}}
% \toprule
%  \makecell[c]{Environments} &  \makecell[c]{MetaCURE}  & \makecell[c]{PEARL} &  \makecell[c]{VariBAD} & \makecell[c]{MAML}  & \makecell[c]{ProMP} & \makecell[c]{RL$^2$}  &\makecell[c] {E-RL$^2$}\\
% \midrule
% \multicolumn{1}{m{0.4\columnwidth}|}{Meta-World Reach} & \makecell[c]{\textbf{0.36$\pm$0.04}} & \makecell[c]{0.16$\pm$0.07} & \makecell[c]{0$\pm$0} & \makecell[c]{0$\pm$0}&  \makecell[c]{0$\pm$0}& \makecell[c] {0$\pm$0}& \makecell[c] {0$\pm$0}\\ 
% %  \hline
% \multicolumn{1}{m{0.4\columnwidth}|}{Meta-World Reach with Wall} &\makecell[c]{\textbf{0.35$\pm$0.11}} & \makecell[c]{0$\pm$0} & \makecell[c]{0.27$\pm$0.04} &\makecell[c]{0$\pm$0} & \makecell[c]{0$\pm$0}&  \makecell[c]{0$\pm$0}& \makecell[c] {0$\pm$0}\\

% \bottomrule
% \end{tabular}
% \end{sc}
% \end{small}
% \end{center}
% %\vskip -0.1in
% \end{table*}

\textbf{Algorithm setup:} MetaCURE is compared against several representative meta-RL algorithms, including PEARL \citep{rakelly2019efficient}, VariBAD \citep{zintgraf2019varibad}, RL$^2$ \citep{duan2016rl}, MAML \citep{finn2017model}, ProMP \citep{rothfuss2019promp}, E-RL$^2$ \citep{Stadie2018SomeCO} and MAME \citep{gurumurthy2020mame}. We also compare with a variant of EPI \citep{zhou2018environment}, which considers both dynamics predictions and reward predictions\footnote{EPI is not trained end-to-end, and we plot its final performance in dash line.}. We use open-source codes provided by the original papers, and performance is averaged over six random seeds.
%MetaCURE is implemented with PyTorch \citep{paszke2019pytorch}.
%The exploitation policy is only performed in the last adaptation episode, and the exploration policy is used for other adaptation episodes. 

\begin{figure*}[h!]
%\vskip 0.1in
\centering
        \subfigure{\includegraphics[width=0.65\columnwidth]{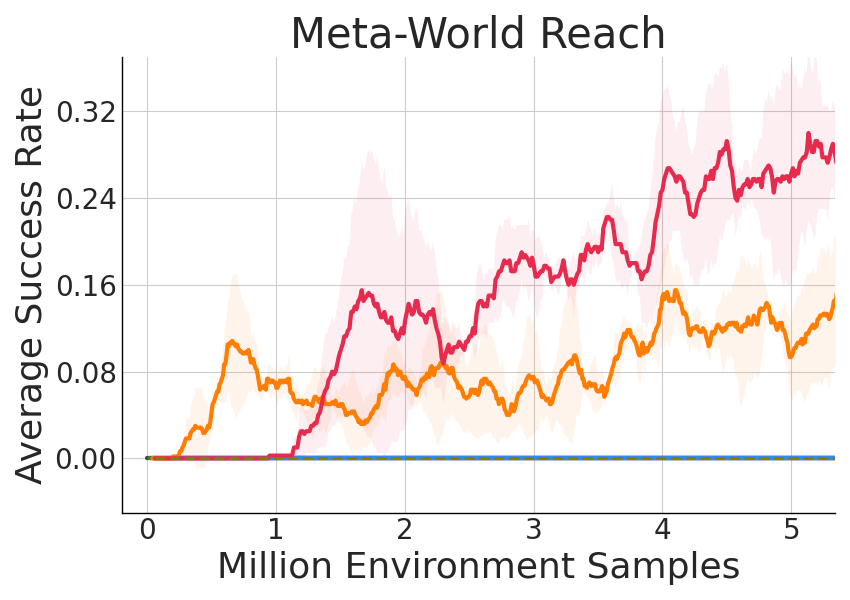}}
        \hspace{0.1\columnwidth}
        \subfigure{\includegraphics[width=0.65\columnwidth]{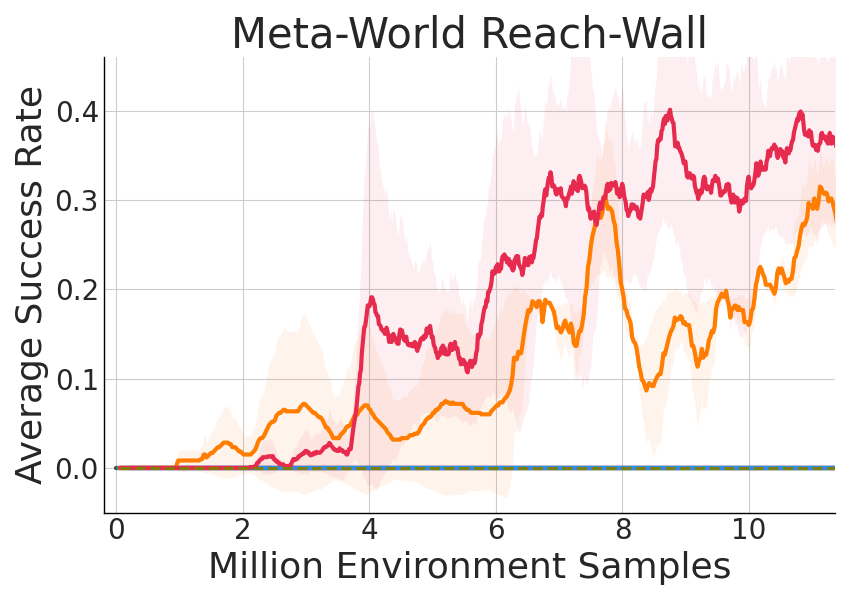}}
        \subfigure{\includegraphics[width=1.35\columnwidth]{novelintr/legend-7.png}}
        % \subfigure{\includegraphics[width=1.9\columnwidth]{novelintr/legend-5.png}}
        \caption{Evaluation of MetaCURE and meta-RL baselines on the challenging sparse-reward Meta-World task sets.
%MetaCURE outperforms baselines algorithms in both final performance and sample efficiency.
}
\label{metaworld}
%\vskip -0.1in
\end{figure*}

\textbf{Results and analyses:} algorithms' performance is evaluated by the last adaptation episode's return averaged over all meta-testing tasks. As shown in Figure \ref{curves}, MetaCURE significantly outperforms baseline algorithms. PEARL fails to achieve satisfactory performance, as it utilizes posterior sampling \citep{thompson1933likelihood,osband2013more} for exploration, which is only optimal in asymptotic performance and may fail to effectively explore within short adaptation. In contrast, MetaCURE learns a separate policy for efficient exploration and achieves superior performance. RL$^2$ performs well in Point-Robot-Sparse, a task set with simple dynamics, but fails in more complex tasks. This is possibly because that RL$^2$ fails to effectively handle task uncertainty in complex sparse-reward tasks. MetaCURE achieves better performance in complex tasks by utilizing probabilistic  task inference to model task uncertainty.
% MAML acquires competitive performance in Walker-Params-Sparse and Hopper-Rand-Params, but fails to learn tasks with different reward functions, as task-specific policies in Walker-Params-Sparse and Hopper-Rand-Params are less distinct, and are easier to adapt with few step gradient updates. VariBAD achieves similar final performance to MetaCURE in Reacher-Goal-Sparse and Point-Robot-Sparse, but underperforms in other tasks. This is because that VariBAD do not optimize for last adaptation rollout's performance, and explore less effectively in adaptation.
% This is probably because gradient-based methods are more sensitive to differences in dynamics. 
The rest of the baselines perform poorly with sparse rewards. As for sample efficiency, MetaCURE and PEARL outperform other methods by realizing off-policy training. Note that MetaCURE and PEARL achieve similar sample efficiency, while MetaCURE learns two policies and PEARL only learns one. This superior sample efficiency is acquired by sharing the task inference component and training data.

\subsection{Adaptation Performance on Meta-World}
\label{res2}
Meta-World \citep{yu2019meta} is a recently proposed challenging evaluation benchmark for meta-RL, including a variety of robot arm control tasks. We evaluate MetaCURE as well as baselines on two Meta-World task sets: Reach and Reach-Wall. To investigate  the efficiency of exploration, we make the rewards sparse, providing non-zero rewards only when the agent succeeds in the task. This setting is extremely hard, as task information is very scarce. Following the original paper \citep{yu2019meta}, we evaluate algorithms by their final success rates, and results are shown in Figure \ref{metaworld}. MetaCURE achieves significantly higher success rates than baselines by achieving efficient exploration. Among the baselines, PEARL is the only algorithm that manages to solve the tasks, but its success rate is significantly lower than MetaCURE.

\subsection{Adaptation Visualization}
\label{res3}
\begin{figure*}[tb]
%\vskip 0.2in
\centering
        \hspace{0.01\columnwidth}
        \subfigure[MetaCURE]{\includegraphics[width=0.76\columnwidth]{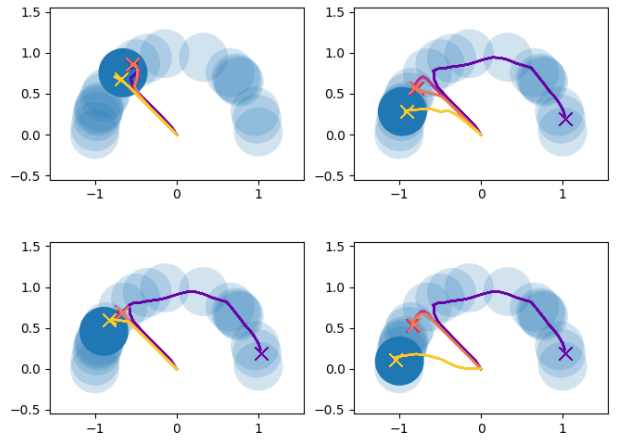}\label{m}}
        \hspace{0.24\columnwidth}
        \subfigure[PEARL]{\includegraphics[width=0.775\columnwidth]{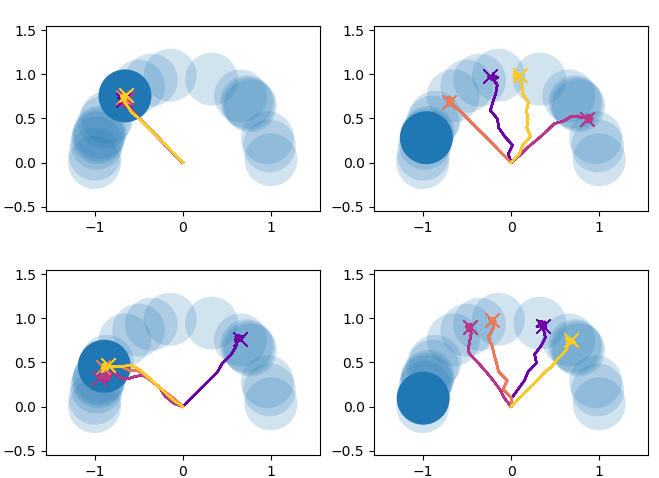}\label{p}}
\caption{Adaptation visualization of (a) MetaCURE and (b) PEARL on Point-Robot-Sparse. The agent is given four episodes to perform adaptation. Purple lines indicate the first adaptation episode's trajectories, while light yellow lines indicate the last adaptation episode's trajectories. Dark blue circles represent rewarding regions for the current task, while light blue circles represent rewarding regions for other meta-testing tasks. MetaCURE achieves more efficient exploration.
% PEARL explores by rolling out exploitation policies. MetaCURE is intrinsically motivated to gather task-specific knowledge, efficiently exploring possible goals and achieving efficient exploration and exploitation behaviors.
}
\label{point-visual}
%\vskip -0.2in
\end{figure*}

\begin{figure*}[t]
%\vskip 0.2in
\centering
        \subfigure{\includegraphics[width=0.85\columnwidth]{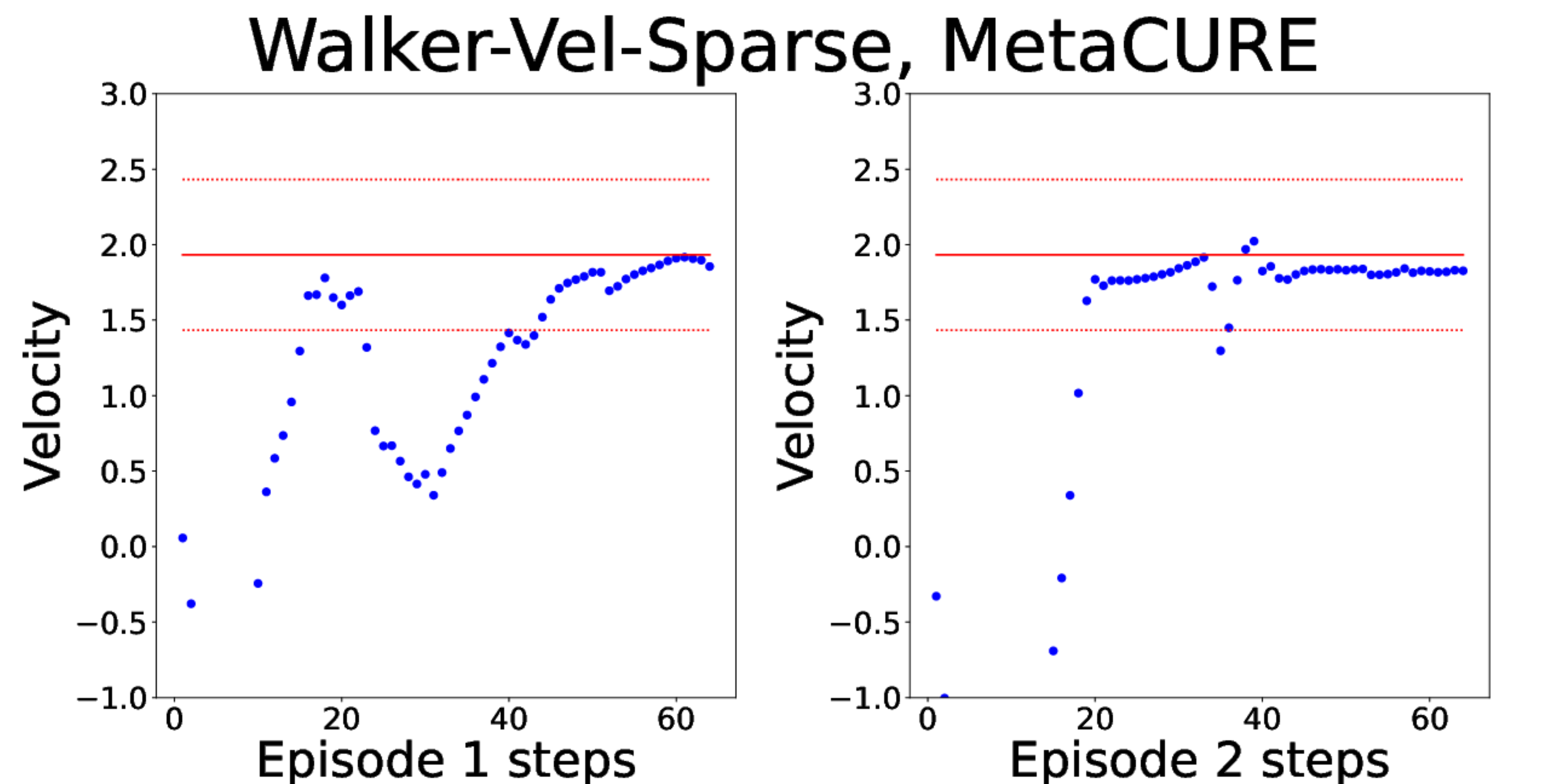}}  
        \hspace{0.12\columnwidth}
        \subfigure{\includegraphics[width=0.85\columnwidth]{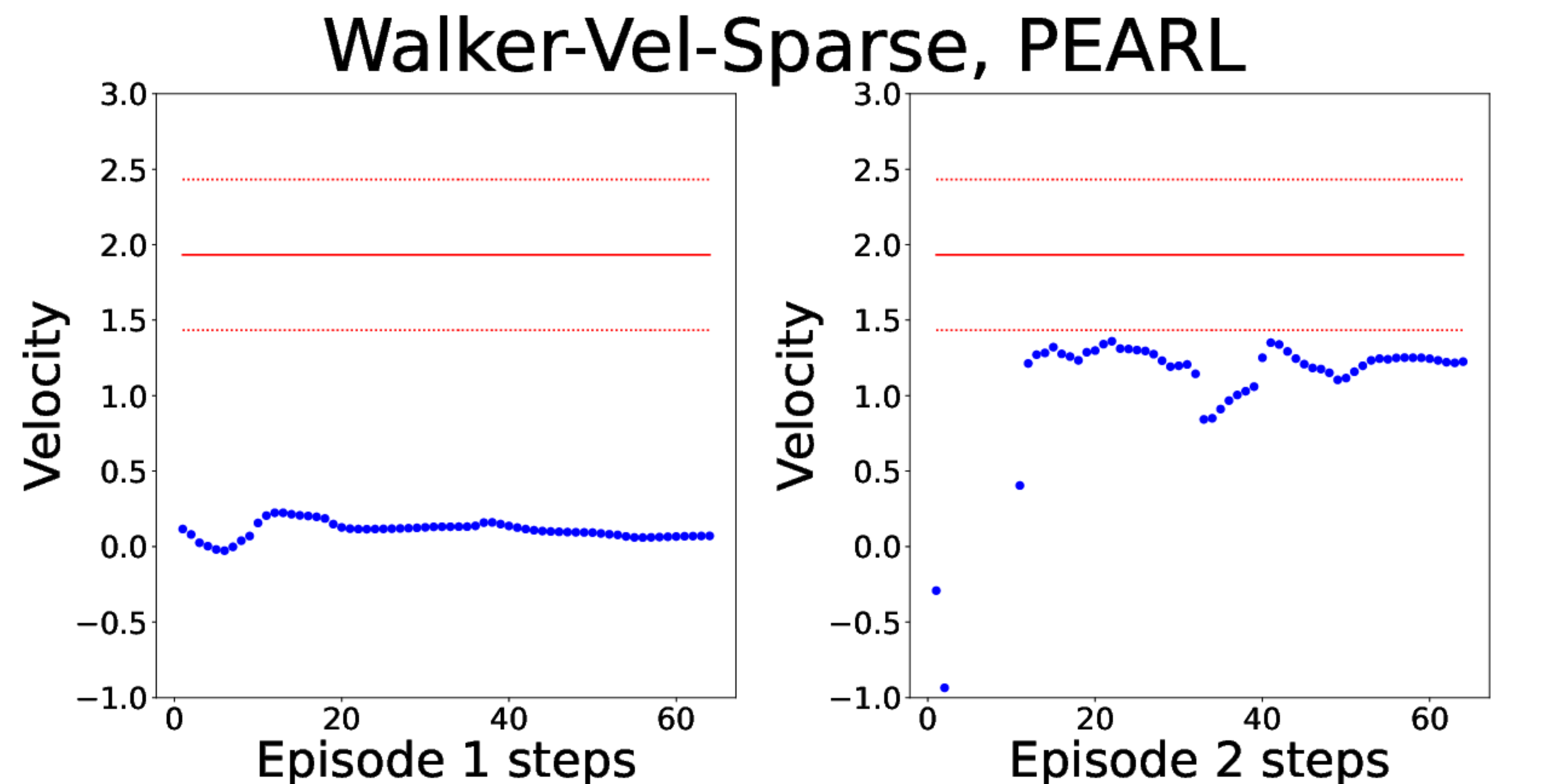}}
\caption{Visualization of MetaCURE and PEARL on Walker-Vel-Sparse. The agent is given two episodes to perform adaptation. The solid red line is the target velocity, and the region bounded by red dash lines represents velocities that get informative rewards. While PEARL tries to keep a certain velocity for an entire episode, MetaCURE first efficiently explores the goal velocity before performing exploitation behaviors.}
\label{walker-visual}
%\vskip -0.2in
\end{figure*}

To prove that MetaCURE learns efficient exploration and exploitation strategies, we visualize MetaCURE's adaptation phase in Point-Robot-Sparse and Walker-Vel-Sparse, as shown in Figure \ref{point-visual} and \ref{walker-visual}, respectively. We compare against PEARL, which explores via posterior sampling. 

In Point-Robot-Sparse (Figure \ref{point-visual}), the agent needs to identify the task by exploring regions where the goal may exist, and then carry out exploitation behaviors. While MetaCURE efficiently explores to identify goal location and then exploits, PEARL only explores a small region every episode, as its policy only performs exploitation behaviors and can not explore effectively.

In Walker-Vel-Sparse (Figure \ref{walker-visual}), the agent needs to identify the goal velocity. MetaCURE covers possible velocities in the first episode with the exploration policy, and then reaches the goal velocity with the exploitation policy in the second episode. In contrast, PEARL only covers a small range of velocities in an episode. Additional visualization results are deferred to Appendix D.

\subsection{Ablation Study}
\label{res4}
This section evaluates the essentiality of MetaCURE's components, including the intrinsic reward, separation of exploitation and exploration policies, and exploration policy using extrinsic rewards.

\begin{figure}[t]
%\vskip 0.2in
\centering
         \subfigure{\includegraphics[width=0.69\columnwidth]{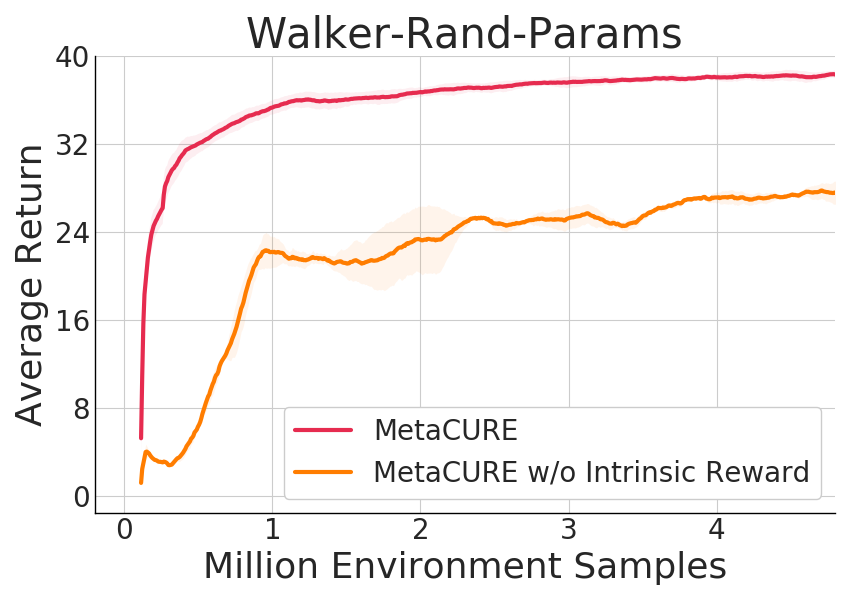}}  

\caption{Ablation study on MetaCURE's intrinsic reward.}
\label{intr2}
\end{figure}

\textbf{Intrinsic reward:} to demonstrate the effectiveness of our intrinsic reward, we test a variant of MetaCURE that ablates the intrinsic reward. As shown in Figure \ref{intr2}, this variant suffers from a massive decrease in performance. This result shows that our intrinsic reward is critical for efficient exploration.

\begin{figure}[t]
%\vskip 0.2in
\centering
         \subfigure{\includegraphics[width=0.69\columnwidth]{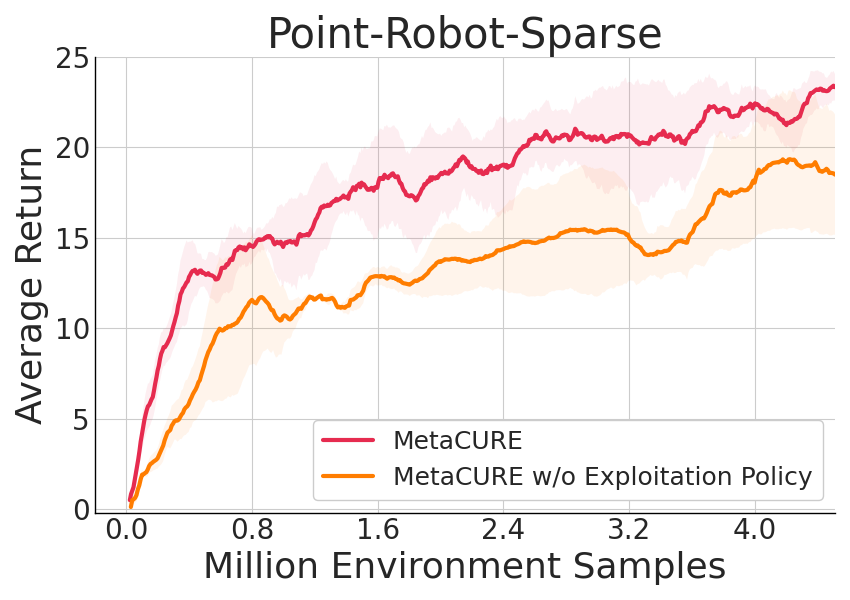}}  
\caption{Ablation study on MetaCURE's exploitation policy.}
\label{explo}
\end{figure}

\textbf{Exploitation policy:} to investigate the effectiveness of learning separate exploration and exploitation policies, we test a variant of MetaCURE which does not obtain an exploitation policy, and utilizes rewards and dynamics prediction as the decoder. As shown in Figure \ref{explo}, on Point-Robot-Sparse, this variant significantly underperforms the original MetaCURE. The exploitation policy is critical for superior performance, as it serves two important purposes: learning unbiased exploitation behaviors, and providing a decoding objective for training the context encoder.

\begin{figure}[t]
%\vskip 0.2in
\centering
         \subfigure{\includegraphics[width=0.69\columnwidth]{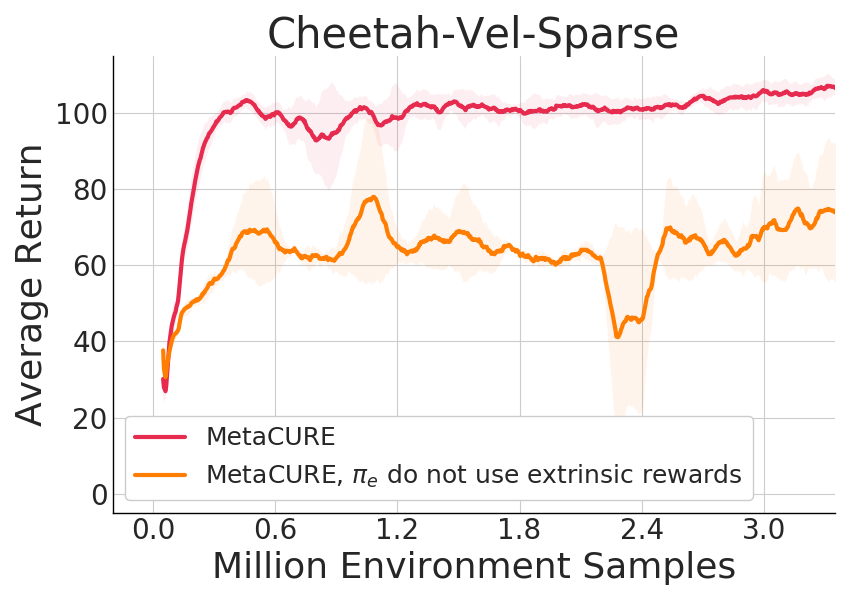}}  
\caption{Ablation study on MetaCURE's exploration policy using extrinsic rewards.}
\label{explore}
\end{figure}

\textbf{Exploration policy using extrinsic rewards:} as shown in Figure \ref{explore}, we find that on Cheetah-Vel-Sparse, by adding extrinsic rewards to the exploration policy's reward function, MetaCURE achieves superior performance compared to the variant that only maximizes intrinsic rewards. This is because that extrinsic reward signals contain rich task information, and can effectively guide exploration. 

Additional ablation studies on hyper-parameters, knowledge and experience sharing as well as a baseline with our intrinsic reward are deferred to Appendix E.

% \textbf{Hyper-parameters:} we conduct ablation studies on MetaCURE's hyper-parameters, as shown in Appendix A.5. Results show that MetaCURE is generally robust to hyper-parameters and does not need particular fine-tuning.

% \textbf{Baseline with intrinsic reward:} we test a variant of VariBAD that uses $r_{exploration}$ for training its policy. As shown in Appendix A.6, this variant does not achieve better performance than the original VariBAD, and still underperforms MetaCURE. This is because that VariBAD does not separate exploration and exploitation policies, and cannot learn intrinsically motivated exploration behaviors and unbiased exploitation behaviors together.

% \textbf{Knowledge and experience sharing:} we test two variants of MetaCURE that ablate either the sharing of task inference component or the sharing of training data. As shown in Appendix A.7, these two variants fail to learn effectively.
%These poor performances are caused by the fact that the intrinsic rewards are hard to diminish during adaptation, and the meta-policies are biased from optimal exploitation behaviors. Meanwhile, MetaCURE achieves superior performance by utilizing an exploiter that maximizes extrinsic return.

%\textbf{Exploration policy structure} We test MetaCURE with exploration policies that only take $s$ as input, ignoring context $c$. This policy structure does not have access to the belief $q_{\phi}(z|c)$ and is confused by the intrinsic reward, which conditions on $c$, resulting in poor performances.

\section{Related Work}
\label{related}

%\textbf{Meta learning} Research on cognitive science shows that children as young as four years old can integrate and make use of prior knowledge in order to master new skills quickly \citep{gopnik2012reconstructing}. The idea of meta-learning \citep{schmidhuber1987evolutionary,naik1992meta,thrun2012learning,finn2017model} is one promising way to equip AIs with the knowledge transferring ability. Meta-learning focuses on extracting and utilizing prior knowledge from a set of tasks to enable fast adaptation on novel unseen tasks. This idea can be applied to all kinds of learning tasks, such as regression, classification, and reinforcement learning.

\textbf{Exploration in meta-RL:} in contrast to meta supervised learning, in meta-RL \citep{schmidhuber1995learning,finn2017model} the agent is not given a task-specific dataset to adapt to, and it must explore the environment to collect useful information. This exploration part is vital for both meta-training and adaptation \citep{schmidhuber1997s,Meta-learning}.  

The problem of exploration policy learning in gradient-based meta-RL is mainly addressed by computing gradients to the pre-update policy's state distribution \citep{Stadie2018SomeCO,rothfuss2019promp}.
%These methods utilize policy injected with random noise for meta-training exploration, which is ineffective and empirically fail with sparse rewards. 
MAESN \citep{gupta2018meta} introduces temporally-extended exploration with latent variables. These methods generally require dense rewards in meta-training. MAME \citep{gurumurthy2020mame} augments MAML \citep{finn2017model} with a separate exploration policy, but its exploration policy learning shares the same objective with exploitation policy learning and is not effective for general sparse-reward tasks. A branch of context-based methods automatically learns to trade-off exploration and exploitation by maximizing average adaptation performance \citep{duan2016rl,zintgraf2019varibad}, and E-RL$^2$ \citep{Stadie2018SomeCO} directly optimizes for the final performance. PEARL \citep{rakelly2019efficient} utilizes posterior sampling \citep{thompson1933likelihood,osband2013more} for exploration. EPI \citep{zhou2018environment} considers the setting of tasks with different dynamics, and introduces intrinsic rewards based on prediction improvement in dynamics. In contrast, we propose an empowerment-driven exploration objective aiming to maximize information gain about the current task, and derive a corresponding intrinsic reward to achieve efficient exploration in both meta-training and adaptation.
%MAESN \citep{gupta2018meta} claims to improve exploration by learning a latent variable that injects structured exploration noises, which is actually also a form of posterior sampling. 


\textbf{Exploration with information-theoretic intrinsic rewards:} encouraging the agent to gain task information is a promising way to facilitate exploration \citep{storck1995reinforcement}. VIME \citep{houthooft2016vime} measures the mutual information between trajectories and the transition function, while EMI \citep{kim2019emi} measures the mutual information between state and action with the next state in consideration, as well as the mutual information between state and next state with the action in consideration, both in the latent space. \citet{sun2011planning} discusses exploration with information gain, but is restricted to planning problems and requires an oracle estimating the posterior. These works focus on traditional RL. Our intrinsic reward maximizes the information gain for task identification during meta-RL's training and adaptation, which is different from previous works.
%To our best knowledge, no previous work has introduced information-theoretic intrinsic rewards into meta-RL.
 
\textbf{Prediction loss as intrinsic reward:} prediction losses can serve as a kind of curiosity-driven intrinsic reward to encourage exploration \citep{schmidhuber1991possibility}, and is widely used in traditional RL.
%Intuitively, high prediction loss implies that the transition has not been fully explored, and visiting these transitions helps the agent to explore.
\citet{oh2015action} directly predicts the image observation, and \citet{stadie2015incentivizing} utilizes prediction loss in the latent space, in order to focus on useful features extracted from observations. To avoid trivial solutions in learning the latent space, \citet{pathak2017curiosity} introduces an inverse model to guide the learning of latent space, only predicting things that the agent can control. \citet{burda2018large} utilizes random neural networks as projections onto the latent space and achieved superior performance on Atari games. Our work focuses on the exploration problem in meta-RL, and utilizes differences of model prediction errors as a means to measure information gain.

\section{Conclusion}

In this paper, to enable efficient meta-RL with sparse rewards, we explicitly model the problem of exploration policy learning, and propose a novel empowerment-driven exploration objective, which aims at maximizing agent's information gain about the current task. We derive a corresponding intrinsic reward from our exploration objective, and a didactic example shows that our intrinsic reward facilitates efficient exploration in both meta-training and adaptation. 
A new off-policy meta-RL algorithm called MetaCURE is also proposed, which incorporates our intrinsic reward and learns separate exploration and exploitation policies. MetaCURE achieves superior performance on various sparse-reward MuJoCo locomotion task sets as well as more difficult sparse-reward Meta-World tasks. 

\section*{Acknowledgements}
This work is supported in part by Science and Technology Innovation 2030 – ``New Generation Artificial Intelligence'' Major Project (No. 2018AAA0100904), and a grant from the Institute of Guo Qiang, Tsinghua University.

\bibliography{example_paper}

\begin{thebibliography}{40}
\providecommand{\natexlab}[1]{#1}
\providecommand{\url}[1]{\texttt{#1}}
\expandafter\ifx\csname urlstyle\endcsname\relax
  \providecommand{\doi}[1]{doi: #1}\else
  \providecommand{\doi}{doi: \begingroup \urlstyle{rm}\Url}\fi

\bibitem[Alemi et~al.(2016)Alemi, Fischer, Dillon, and Murphy]{alemi2016deep}
Alemi, A.~A., Fischer, I., Dillon, J.~V., and Murphy, K.
\newblock Deep variational information bottleneck.
\newblock \emph{arXiv preprint arXiv:1612.00410}, 2016.

\bibitem[Babaeizadeh et~al.(2018)Babaeizadeh, Finn, Erhan, Campbell, and
  Levine]{babaeizadeh2017stochastic}
Babaeizadeh, M., Finn, C., Erhan, D., Campbell, R.~H., and Levine, S.
\newblock Stochastic variational video prediction.
\newblock In \emph{International Conference on Learning Representations}, 2018.

\bibitem[Burda et~al.(2018{\natexlab{a}})Burda, Edwards, Pathak, Storkey,
  Darrell, and Efros]{burda2018large}
Burda, Y., Edwards, H., Pathak, D., Storkey, A., Darrell, T., and Efros, A.~A.
\newblock Large-scale study of curiosity-driven learning.
\newblock In \emph{International Conference on Learning Representations},
  2018{\natexlab{a}}.

\bibitem[Burda et~al.(2018{\natexlab{b}})Burda, Edwards, Storkey, and
  Klimov]{burda2018exploration}
Burda, Y., Edwards, H., Storkey, A., and Klimov, O.
\newblock Exploration by random network distillation.
\newblock In \emph{International Conference on Learning Representations},
  2018{\natexlab{b}}.

\bibitem[Chung et~al.(2015)Chung, Kastner, Dinh, Goel, Courville, and
  Bengio]{chung2015recurrent}
Chung, J., Kastner, K., Dinh, L., Goel, K., Courville, A., and Bengio, Y.
\newblock A recurrent latent variable model for sequential data.
\newblock In \emph{Proceedings of the 28th International Conference on Neural
  Information Processing Systems-Volume 2}, pp.\  2980--2988, 2015.

\bibitem[Duan et~al.(2016)Duan, Schulman, Chen, Bartlett, Sutskever, and
  Abbeel]{duan2016rl}
Duan, Y., Schulman, J., Chen, X., Bartlett, P.~L., Sutskever, I., and Abbeel,
  P.
\newblock Rl $^{2}$: Fast reinforcement learning via slow reinforcement
  learning.
\newblock \emph{arXiv preprint arXiv:1611.02779}, 2016.

\bibitem[Finn \& Levine(2019)Finn and Levine]{Meta-learning}
Finn, C. and Levine, S.
\newblock Meta-learning: from few-shot learning to rapid reinforcement
  learning.
\newblock \url{https://sites.google.com/view/icml19metalearning}, 2019.

\bibitem[Finn et~al.(2017)Finn, Abbeel, and Levine]{finn2017model}
Finn, C., Abbeel, P., and Levine, S.
\newblock Model-agnostic meta-learning for fast adaptation of deep networks.
\newblock In \emph{Proceedings of the 34th International Conference on Machine
  Learning-Volume 70}, pp.\  1126--1135. JMLR. org, 2017.

\bibitem[Gupta et~al.(2018)Gupta, Mendonca, Liu, Abbeel, and
  Levine]{gupta2018meta}
Gupta, A., Mendonca, R., Liu, Y., Abbeel, P., and Levine, S.
\newblock Meta-reinforcement learning of structured exploration strategies.
\newblock In \emph{Advances in Neural Information Processing Systems}, pp.\
  5302--5311, 2018.

\bibitem[Gurumurthy et~al.(2020)Gurumurthy, Kumar, and
  Sycara]{gurumurthy2020mame}
Gurumurthy, S., Kumar, S., and Sycara, K.
\newblock Mame: Model-agnostic meta-exploration.
\newblock In \emph{Conference on Robot Learning}, pp.\  910--922. PMLR, 2020.

\bibitem[Haarnoja et~al.(2018)Haarnoja, Zhou, Abbeel, and
  Levine]{haarnoja2018soft}
Haarnoja, T., Zhou, A., Abbeel, P., and Levine, S.
\newblock Soft actor-critic: Off-policy maximum entropy deep reinforcement
  learning with a stochastic actor.
\newblock In \emph{International Conference on Machine Learning}, pp.\
  1856--1865, 2018.

\bibitem[Hafner et~al.(2019)Hafner, Lillicrap, Ba, and
  Norouzi]{hafner2019dream}
Hafner, D., Lillicrap, T., Ba, J., and Norouzi, M.
\newblock Dream to control: Learning behaviors by latent imagination.
\newblock In \emph{International Conference on Learning Representations}, 2019.

\bibitem[Hessel et~al.(2018)Hessel, Modayil, Van~Hasselt, Schaul, Ostrovski,
  Dabney, Horgan, Piot, Azar, and Silver]{hessel2018rainbow}
Hessel, M., Modayil, J., Van~Hasselt, H., Schaul, T., Ostrovski, G., Dabney,
  W., Horgan, D., Piot, B., Azar, M., and Silver, D.
\newblock Rainbow: Combining improvements in deep reinforcement learning.
\newblock In \emph{Thirty-Second AAAI Conference on Artificial Intelligence},
  2018.

\bibitem[Houthooft et~al.(2016)Houthooft, Chen, Duan, Schulman, De~Turck, and
  Abbeel]{houthooft2016vime}
Houthooft, R., Chen, X., Duan, Y., Schulman, J., De~Turck, F., and Abbeel, P.
\newblock Vime: Variational information maximizing exploration.
\newblock In \emph{Advances in Neural Information Processing Systems}, pp.\
  1109--1117, 2016.

\bibitem[Humplik et~al.(2019)Humplik, Galashov, Hasenclever, Ortega, Teh, and
  Heess]{humplik2019meta}
Humplik, J., Galashov, A., Hasenclever, L., Ortega, P.~A., Teh, Y.~W., and
  Heess, N.
\newblock Meta reinforcement learning as task inference.
\newblock \emph{arXiv preprint arXiv:1905.06424}, 2019.

\bibitem[Kim et~al.(2019)Kim, Kim, Jeong, Levine, and Song]{kim2019emi}
Kim, H., Kim, J., Jeong, Y., Levine, S., and Song, H.~O.
\newblock Emi: Exploration with mutual information.
\newblock In \emph{International Conference on Machine Learning}, pp.\
  3360--3369, 2019.

\bibitem[Kingma \& Welling(2013)Kingma and Welling]{kingma2013auto}
Kingma, D.~P. and Welling, M.
\newblock Auto-encoding variational bayes.
\newblock \emph{arXiv preprint arXiv:1312.6114}, 2013.

\bibitem[Lee et~al.(2020)Lee, Seo, Lee, Lee, and Shin]{lee2020context}
Lee, K., Seo, Y., Lee, S., Lee, H., and Shin, J.
\newblock Context-aware dynamics model for generalization in model-based
  reinforcement learning.
\newblock In \emph{International Conference on Machine Learning}, pp.\
  5757--5766. PMLR, 2020.

\bibitem[Leike et~al.(2016)Leike, Lattimore, Orseau, and
  Hutter]{leike2016thompson}
Leike, J., Lattimore, T., Orseau, L., and Hutter, M.
\newblock Thompson sampling is asymptotically optimal in general environments.
\newblock In \emph{Proceedings of the Thirty-Second Conference on Uncertainty
  in Artificial Intelligence}, pp.\  417--426, 2016.

\bibitem[Mishra et~al.(2018)Mishra, Rohaninejad, Chen, and
  Abbeel]{mishra2017simple}
Mishra, N., Rohaninejad, M., Chen, X., and Abbeel, P.
\newblock A simple neural attentive meta-learner.
\newblock In \emph{International Conference on Learning Representations}, 2018.

\bibitem[Oh et~al.(2015)Oh, Guo, Lee, Lewis, and Singh]{oh2015action}
Oh, J., Guo, X., Lee, H., Lewis, R.~L., and Singh, S.
\newblock Action-conditional video prediction using deep networks in atari
  games.
\newblock In \emph{Advances in neural information processing systems}, pp.\
  2863--2871, 2015.

\bibitem[Osband et~al.(2013)Osband, Russo, and Van~Roy]{osband2013more}
Osband, I., Russo, D., and Van~Roy, B.
\newblock (more) efficient reinforcement learning via posterior sampling.
\newblock In \emph{Advances in Neural Information Processing Systems}, pp.\
  3003--3011, 2013.

\bibitem[Paszke et~al.(2019)Paszke, Gross, Massa, Lerer, Bradbury, Chanan,
  Killeen, Lin, Gimelshein, Antiga, et~al.]{paszke2019pytorch}
Paszke, A., Gross, S., Massa, F., Lerer, A., Bradbury, J., Chanan, G., Killeen,
  T., Lin, Z., Gimelshein, N., Antiga, L., et~al.
\newblock Pytorch: An imperative style, high-performance deep learning library.
\newblock In \emph{Advances in Neural Information Processing Systems}, pp.\
  8024--8035, 2019.

\bibitem[Pathak et~al.(2017)Pathak, Agrawal, Efros, and
  Darrell]{pathak2017curiosity}
Pathak, D., Agrawal, P., Efros, A.~A., and Darrell, T.
\newblock Curiosity-driven exploration by self-supervised prediction.
\newblock In \emph{Proceedings of the IEEE Conference on Computer Vision and
  Pattern Recognition Workshops}, pp.\  16--17, 2017.

\bibitem[Rakelly et~al.(2019)Rakelly, Zhou, Finn, Levine, and
  Quillen]{rakelly2019efficient}
Rakelly, K., Zhou, A., Finn, C., Levine, S., and Quillen, D.
\newblock Efficient off-policy meta-reinforcement learning via probabilistic
  context variables.
\newblock In \emph{International Conference on Machine Learning}, pp.\
  5331--5340, 2019.

\bibitem[Rothfuss et~al.(2019)Rothfuss, Lee, Clavera, Asfour, Abbeel,
  Shingarey, Kaul, Asfour, Athanasios, Zhou, et~al.]{rothfuss2019promp}
Rothfuss, J., Lee, D., Clavera, I., Asfour, T., Abbeel, P., Shingarey, D.,
  Kaul, L., Asfour, T., Athanasios, C.~D., Zhou, Y., et~al.
\newblock Promp: Proximal meta-policy search.
\newblock In \emph{International Conference on Learning Representations},
  volume~3, pp.\  4007--4014, 2019.

\bibitem[Schmidhuber(1991)]{schmidhuber1991possibility}
Schmidhuber, J.
\newblock A possibility for implementing curiosity and boredom in
  model-building neural controllers.
\newblock In \emph{Proc. of the international conference on simulation of
  adaptive behavior: From animals to animats}, pp.\  222--227, 1991.

\bibitem[Schmidhuber(1995)]{schmidhuber1995learning}
Schmidhuber, J.
\newblock On learning how to learn learning strategies.
\newblock 1995.

\bibitem[Schmidhuber(1997)]{schmidhuber1997s}
Schmidhuber, J.
\newblock What's interesting?
\newblock 1997.

\bibitem[Stadie et~al.(2015)Stadie, Levine, and
  Abbeel]{stadie2015incentivizing}
Stadie, B.~C., Levine, S., and Abbeel, P.
\newblock Incentivizing exploration in reinforcement learning with deep
  predictive models.
\newblock \emph{arXiv preprint arXiv:1507.00814}, 2015.

\bibitem[Stadie et~al.(2018)Stadie, Yang, Houthooft, Chen, Duan, Wu, Abbeel,
  and Sutskever]{Stadie2018SomeCO}
Stadie, B.~C., Yang, G., Houthooft, R., Chen, X., Duan, Y., Wu, Y., Abbeel, P.,
  and Sutskever, I.
\newblock Some considerations on learning to explore via meta-reinforcement
  learning.
\newblock \emph{ArXiv}, abs/1803.01118, 2018.

\bibitem[Storck et~al.(1995)Storck, Hochreiter, and
  Schmidhuber]{storck1995reinforcement}
Storck, J., Hochreiter, S., and Schmidhuber, J.
\newblock Reinforcement driven information acquisition in non-deterministic
  environments.
\newblock In \emph{Proceedings of the international conference on artificial
  neural networks, Paris}, volume~2, pp.\  159--164. Citeseer, 1995.

\bibitem[Sun et~al.(2011)Sun, Gomez, and Schmidhuber]{sun2011planning}
Sun, Y., Gomez, F., and Schmidhuber, J.
\newblock Planning to be surprised: Optimal bayesian exploration in dynamic
  environments.
\newblock In \emph{International Conference on Artificial General
  Intelligence}, pp.\  41--51. Springer, 2011.

\bibitem[Sutton \& Barto(2018)Sutton and Barto]{sutton2018reinforcement}
Sutton, R.~S. and Barto, A.~G.
\newblock \emph{Reinforcement learning: An introduction}.
\newblock MIT press, 2018.

\bibitem[Thompson(1933)]{thompson1933likelihood}
Thompson, W.~R.
\newblock On the likelihood that one unknown probability exceeds another in
  view of the evidence of two samples.
\newblock \emph{Biometrika}, 25\penalty0 (3/4):\penalty0 285--294, 1933.

\bibitem[Todorov et~al.(2012)Todorov, Erez, and Tassa]{todorov2012Mujoco}
Todorov, E., Erez, T., and Tassa, Y.
\newblock Mujoco: A physics engine for model-based control.
\newblock In \emph{2012 IEEE/RSJ International Conference on Intelligent Robots
  and Systems}, pp.\  5026--5033. IEEE, 2012.

\bibitem[Vinyals et~al.(2019)Vinyals, Babuschkin, Czarnecki, Mathieu, Dudzik,
  Chung, Choi, Powell, Ewalds, Georgiev, et~al.]{vinyals2019grandmaster}
Vinyals, O., Babuschkin, I., Czarnecki, W.~M., Mathieu, M., Dudzik, A., Chung,
  J., Choi, D.~H., Powell, R., Ewalds, T., Georgiev, P., et~al.
\newblock Grandmaster level in starcraft ii using multi-agent reinforcement
  learning.
\newblock \emph{Nature}, pp.\  1--5, 2019.

\bibitem[Yu et~al.(2020)Yu, Quillen, He, Julian, Hausman, Finn, and
  Levine]{yu2019meta}
Yu, T., Quillen, D., He, Z., Julian, R., Hausman, K., Finn, C., and Levine, S.
\newblock Meta-world: A benchmark and evaluation for multi-task and meta
  reinforcement learning.
\newblock In \emph{Conference on Robot Learning}, pp.\  1094--1100. PMLR, 2020.

\bibitem[Zhou et~al.(2018)Zhou, Pinto, and Gupta]{zhou2018environment}
Zhou, W., Pinto, L., and Gupta, A.
\newblock Environment probing interaction policies.
\newblock In \emph{International Conference on Learning Representations}, 2018.

\bibitem[Zintgraf et~al.(2019)Zintgraf, Shiarlis, Igl, Schulze, Gal, Hofmann,
  and Whiteson]{zintgraf2019varibad}
Zintgraf, L., Shiarlis, K., Igl, M., Schulze, S., Gal, Y., Hofmann, K., and
  Whiteson, S.
\newblock Varibad: A very good method for bayes-adaptive deep rl via
  meta-learning.
\newblock In \emph{International Conference on Learning Representations}, 2019.

\end{thebibliography}
\bibliographystyle{icml2021}

% \appendix
% \section{Appendix}
% You may include other additional sections here.
% \end{document}
\appendix
\onecolumn
\appendix
\section{Proofs for Section 3.1}
\label{proof1}
% \emph{Proof.}
\begin{align}
    &\mathcal{J}^{\pi_e}(C_{:H},\mathcal{K}) \notag \\
    &=I^{\pi_e}(C_{:H};\mathcal{K}) \notag \\
        &=\mathbb{E}_{{(c_{:H},\kappa)}\sim (C_{:H},\mathcal{K})}\left[\log\frac{ p^{\pi_e}(c_{:H}|\kappa)}{p^{\pi_e}(c_{:H})}\right]\\
        &=\mathbb{E}_{{(c_{:H},\kappa)}\sim (C_{:H},\mathcal{K})} \left[\log \frac{p_0^{\kappa}(s_0)}{p_0(s_0)}+\sum_{t=0}^{H-1}\log \frac{p^{\pi_e}(c_t|c_{:t},\kappa)}{p^{\pi_e}(c_t|c_{:t})} \right] \\
         &=\mathbb{E}_{{(c_{:H},\kappa)}\sim (C_{:H},\mathcal{K})}  \left[\log \frac{p_0^{\kappa}(s_0)}{p_0(s_0)}+\sum_{t=0}^{H-1}\log \frac{p^{\pi_e}(a_t,r_t,s_{t+1}|c_{:t},\kappa)}{p^{\pi_e}(a_t,r_t,s_{t+1}|c_{:t})} \right] \label{proof3}\\
        &=\mathbb{E}_{{(c_{:H},\kappa)}\sim (C_{:H},\mathcal{K})} \left[\log \frac{p_0^{\kappa}(s_0)}{p_0(s_0)}+\sum_{t=0}^{H-1}\log \frac{p^{\pi_e}(a_t|c_{:t},\kappa)p(r_{t},s_{t+1}|{\kappa},c_{:t},a_{t})}{p^{\pi_e}(a_t|c_{:t})p(r_{t},s_{t+1}|c_{:t},a_{t})} \right] \\
        &=\mathbb{E}_{{(c_{:H},\kappa)}\sim (C_{:H},\mathcal{K})}\left[\sum_{t=0}^{H-1}\log \frac{p(r_{t},s_{t+1}|{\kappa},s_{t},a_{t})}{p(r_{t},s_{t+1}|c_{:t},a_{t})} \right]+\textit{Const},
    \label{1}
\end{align}
 where $p_0(s_0)=\mathbb{E}_{\kappa \sim \mathcal{K}}\left[p_0^{\kappa}\left(s_0\right)\right]$ is the marginal distribution of initial state $s_0$. Eq. \eqref{proof3} holds as $c_t=(s_t,a_t,r_t,s_{t+1})$ and $s_t$ is contained in $c_{:t}$. $\textit{Const}=\mathbb{E}_{{(c_{:H},\kappa)}\sim (C_{:H},\mathcal{K})} \left[\log \frac{p_0^{\kappa}(s_0)}{p_0(s_0)}\right]$ is a constant that does not change with the policy $\pi_e$.

\section{Additional Implementation Details}
\label{addition}
\textbf{Encoder structure:} the context encoder should be able to effectively extract task information from experience sequences. In practice, we design a context encoder using temporal convolution and soft attention, similar to SNAIL \citep{mishra2017simple}.

\begin{figure*}[ht]
\centering
         
        \subfigure{\includegraphics[width=0.45\columnwidth]{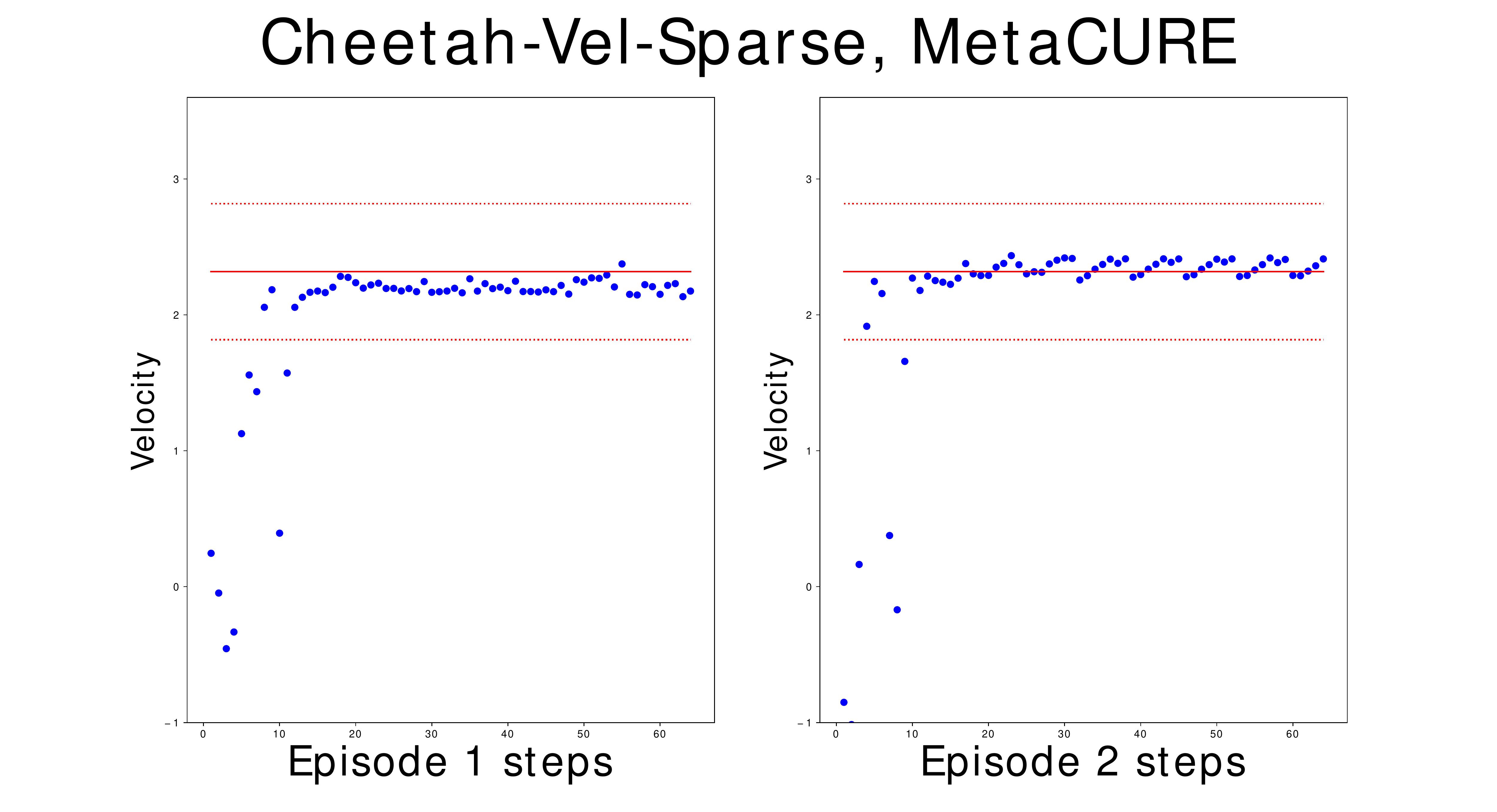}}  
        \subfigure{\includegraphics[width=0.48\columnwidth]{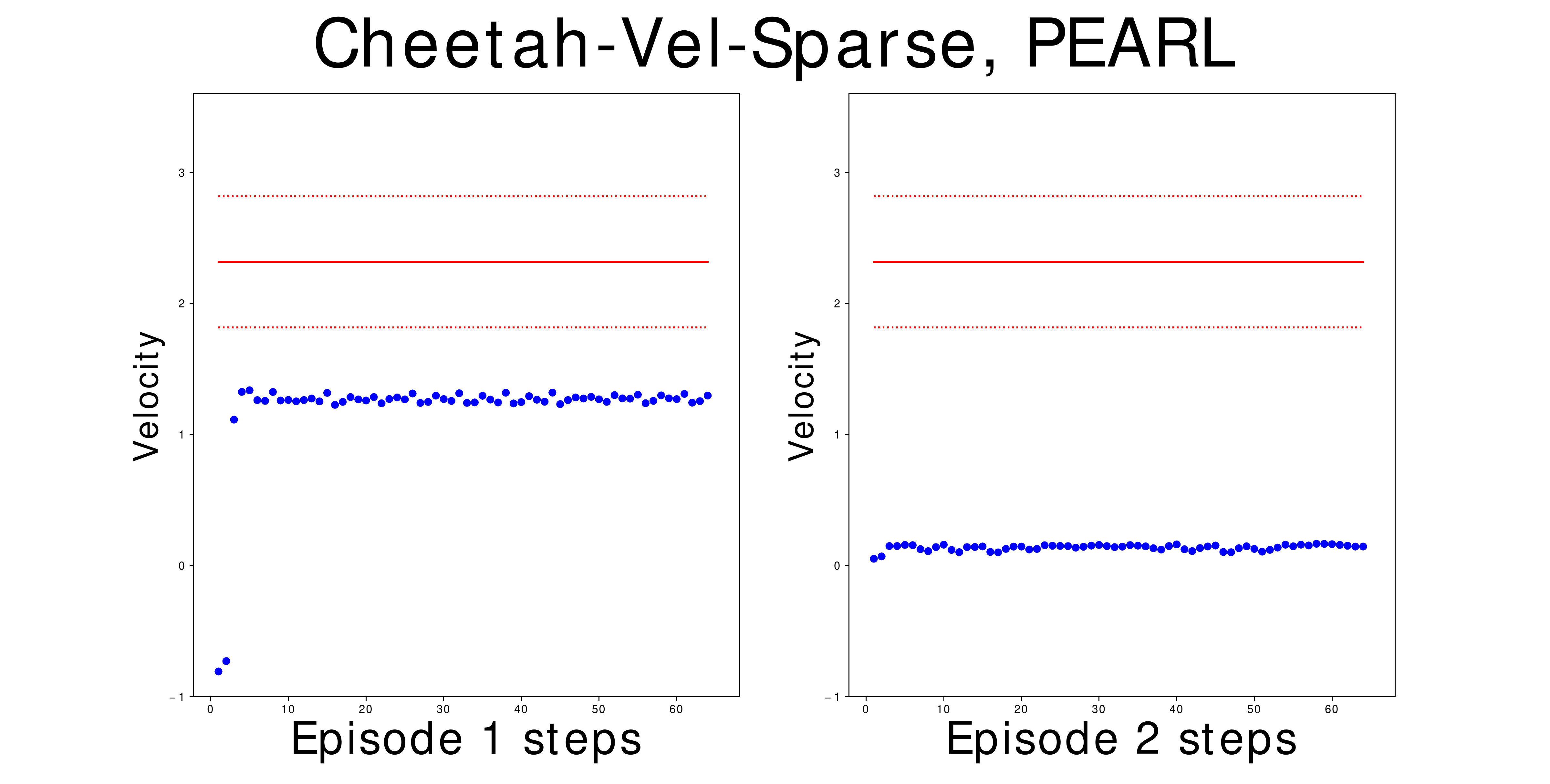}} 
\caption{Visualization of MetaCURE and PEARL on Cheetah-Vel-Sparse. While MetaCURE efficiently explores possible goal velocities in the first episode and exploits in the second episode, PEARL fails to effectively explore.
}
\label{add-visual}
\end{figure*}

%  \textbf{Estimating $L_{pred}$ and $L_{pred}^{task}$:} the log probability is hard to estimate if the rewards and states are continuous. We follow the common approach of utilizing L2 distances as an approximation of the negative log probability \citep{chung2015recurrent,babaeizadeh2017stochastic}:
% \begin{equation}
% \begin{split}
%   &L_{pred}(c_{:t+1})\approx \|r_t-\hat{r}_t^{pred}(c_{:t},a_t)\|_2^2\\
%   &\ \ \ \ \ \ \ \ \ \ \ \ \ \ \ \ \ \ \ \ \ \ \ \ \ \ \ \ \ \ \ \ \ \ \ \ \  +\|s_{t+1}-\hat{s}_{t+1}^{pred}(c_{:t},a_t)\|_2^2 \\
%   &L_{pred}^{task}(\kappa,c_t)\approx \|r_t-\tilde{r}_t^{pred}(\kappa,s_t,a_t)\|_2^2\\
%   & \ \ \ \ \ \ \ \ \ \ \ \ \ \ \ \ \ \ \ \ \ \ \ \ \ \ \ \ \ \ \ \ \ \ \ \ \  +\|s_{t+1}-\tilde{s}_{t+1}^{pred}(\kappa,s_t,a_t)\|_2^2,
% \end{split}
% \label{intrinsicr2}
% \end{equation}{}
% where $\hat{r}_t^{pred}$ and $\hat{s}_{t+1}^{pred}$ are reward and transition predicted by the Meta-Predictor, while $\tilde{r}_t^{pred}$ and $\tilde{s}_{t+1}^{pred}$ are predicted by the Task-Predictor.

\textbf{Other details:} MetaCURE is implemented with PyTorch \citep{paszke2019pytorch}. Generally it takes about 12-40h to converge on the MuJoCo task sets, and can be further accelerated with parallel sampling. Hyper-parameters are selected by a simple grid search.

\section{Environment and Hyper-parameter Settings}
\label{appen-hyper}
In this subsection, we provide detailed settings of reward functions and hyper-parameters, which are shown in Table \ref{env setting} and \ref{hyperparameter}. All tasks obtain sparse reward functions, providing zero rewards if the agent is outside the range of goals. The agent is additionally penalized with control costs.

In the Meta-World tasks, the agent gets non-zero rewards only if it ``successes'' in the task, which is given as a binary signal by the environment.

\section{Visualizations}
We show additional visualization results on Cheetah-Vel-Sparse, as shown in Figure \ref{add-visual}.
\label{visualizations}

\section{Ablation Studies}
\subsection{Ablation Study: Hyper-parameters}
\label{hyper}
We test MetaCURE with different hyper-parameters on Cheetah-Vel-Sparse. As shown in Table \ref{hyper-table}, MetaCURE is generally robust to the choice of hyper-parameters.

\begin{figure*}[ht]
\centering
         
        \subfigure[]{\includegraphics[width=0.45\columnwidth]{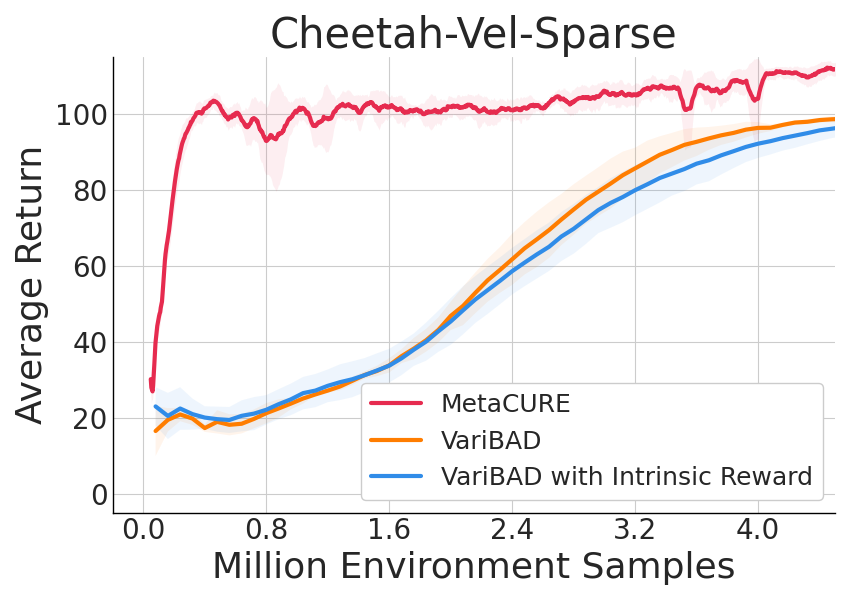}\label{vari2}}  
        \hspace{0.06\columnwidth}
        \subfigure[]{\includegraphics[width=0.45\columnwidth]{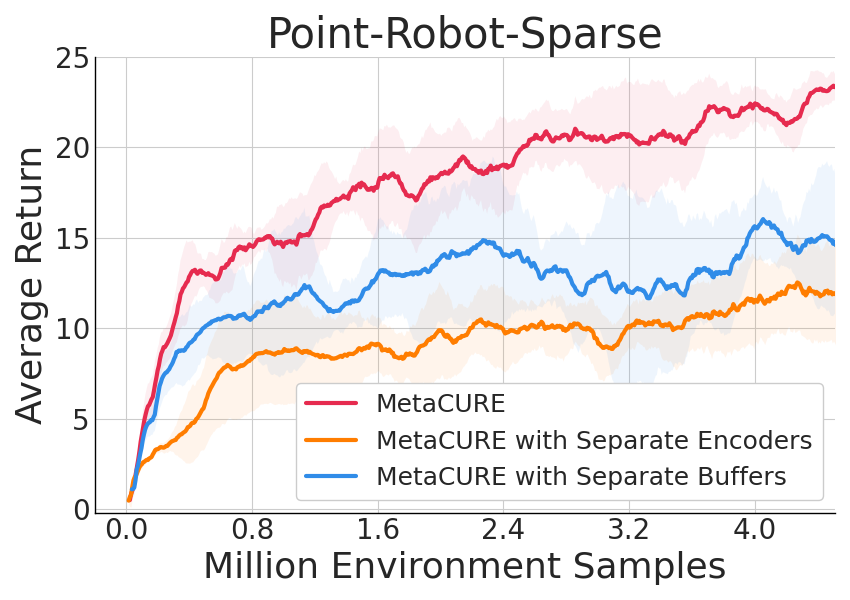}\label{share}} 
\caption{(a) A VariBAD variant that uses our intrinsic reward does not improve in performance. VariBAD does not separate exploration and exploitation, and fails to achieve satisfactory performance with our intrinsic reward. (b) Ablation study on MetaCURE's knowledge and data sharing. Sharing the task inference component and the replay buffer greatly improves MetaCURE's learning efficiency.
}
\end{figure*}

% \subsection{Ablation Study: Intrinsic Reward}
% As shown in Figure \ref{intr2}, our intrinsic reward is essential for efficient learning and superior performance, as it contributes to efficient exploration in both meta-training and adaptation. 
% \label{intrinsic ablation}
% % \begin{figure}[tb]
% % %\vskip 0.2in
% % \centering
% %          \subfigure{\includegraphics[width=0.85\columnwidth]{novelintr/intr-abl.png}}  

% % \caption{Ablation study on MetaCURE's intrinsic reward. Both $L_{pred}$ and $L_{pred}^{task}$ are vital for MetaCURE's performance.}
% % \label{intr2}
% % \end{figure}
% \begin{figure}[tb]
% %\vskip 0.2in
% \centering
%          \subfigure{\includegraphics[width=0.85\columnwidth]{novelintr/wparams-abl.png}}  

% \caption{Ablation study on MetaCURE's intrinsic reward.}
% \label{intr2}
% \end{figure}

\subsection{Ablation Study: Baseline with Intrinsic Reward}
We test a variant of VariBAD \citep{zintgraf2019varibad} that uses $r_{exploration}$ for training its policy. As shown in Figure \ref{vari2}, directly combining our intrinsic reward with VariBAD does not lead to satisfactory performance, and still underperforms MetaCURE. This is because that VariBAD does not separate exploration and exploitation policies, and does not support learning intrinsically motivated exploration behaviors and unbiased exploitation behaviors together.
\label{baseline ablation}

\subsection{Ablation Study: Knowledge and Data Sharing}
To show the effect of sharing task encoder and buffer, we test two variants of MetaCURE on Point-Robot-Sparse: one uses separate encoders for the exploration and exploitation policy, and the other one uses separate buffers for training the policies, as shown in Figure \ref{share}. Both variants suffer from a great decrease in learning efficiency, as data and knowledge are not utilized effectively.
\label{share ablation}
% \begin{figure}[tb]
% %\vskip 0.2in
% \centering
%          \subfigure{\includegraphics[width=0.4\columnwidth]{novelintr/enc-abl1.png}}  

% \caption{Ablation study on MetaCURE's knowledge and data sharing. Sharing the task inference component and the replay buffer greatly improves MetaCURE's learning efficiency.}
% \label{share}
% \end{figure}
\begin{table*}[tb]
\caption{Adaptation length and goal settings for environments used for evaluation. Goals are uniformly distributed in \emph{Goal range} and non-zero informative reward is provided only when the distance between the agent's position/speed and the goal is smaller than \emph{Goal radius}. As for Meta-World Reach and Meta-World Reach-Wall, the goal range and goal radius follow the settings in the original paper \citep{yu2019meta}.}
\label{env setting}
\begin{center}
\begin{small}
\begin{tabular}{|c|c|c|c|c|c|}
\hline
Environment             & \# of adaptation episodes & \begin{tabular}[c]{@{}c@{}}Max steps\\ per episode\end{tabular} & Goal type & Goal range                                                            & Goal radius \\ \hline
Cheetah-Vel-Sparse      & 2                         & 64                                                              & Velocity  & {[}0,3{]}                                                             & 0.5         \\ \hline
Walker-Vel-Sparse       & 2                         & 64                                                              & Velocity  & {[}0,2{]}                                                             & 0.5         \\ \hline
Reacher-Goal-Sparse     & 2                         & 64                                                              & Position  & \begin{tabular}[c]{@{}c@{}}Semicircle with\\ radius 0.25\end{tabular} & 0.09        \\ \hline
Point-Robot-Sparse      & 4                         & 32                                                              & Position  & \begin{tabular}[c]{@{}c@{}}Semicircle with\\ radius 1\end{tabular}    & 0.3         \\ \hline
Walker-Rand-Params      & 4                         & 64                                                              & Velocity  & 1.5                                                                   & 0.5         \\ \hline
Hopper-Rand-Params      & 4                         & 64                                                              & Velocity  & 1.5                                                                   & 0.5         \\ \hline
Meta-World   Reach      & 4                         & 150                                                             & Position  & /                                                                     & /           \\ \hline
Meta-World   Reach-Wall & 4                         & 150                                                             & Position  & /                                                                     & /           \\ \hline
\end{tabular}
\end{small}
\end{center}
\end{table*}

\begin{table*}[tb]
\caption{Hyperparameter settings for MetaCURE in different environments.}
\label{hyperparameter}
%\vskip 0.15in
\begin{center}
\begin{small}
\begin{tabular}{|c|c|c|c|c|c|}
\hline
Environment           & Latent size & $\beta$ & $\lambda$ & Batch size & Learning rate \\ \hline
Cheetah-Vel-Sparse    & 5           & 0.1                  & 5                    & 64         & 3e-4         \\ \hline
Walker-Vel-Sparse     & 5           & 0.1                    & 5                    & 64         & 3e-4         \\ \hline
Reacher-Goal-Sparse   & 5           & 1                    & 1                      & 64         & 3e-4         \\ \hline
Point-Robot-Sparse    & 5           & 1                    & 0.3                      & 96         & 3e-4         \\ \hline
Walker-Rand-Params    & 5           & 1                    & 5                      & 256        & 3e-4         \\ \hline
Hopper-Rand-Params    & 5           & 1                    & 5                      & 256        & 3e-4         \\ \hline
Meta-World Reach      & 5           & 1                    & 0.3                      & 512        & 1e-4         \\ \hline
Meta-World Reach-Wall & 5           & 1                    & 0.3                      & 512        & 1e-4         \\ \hline
\end{tabular}
\end{small}
\end{center}
%\vskip -0.1in
\end{table*}
\begin{table}[t!]
\caption{MetaCURE's hyper-parameter ablation studies on the Cheetah-Vel-Sparse task set.}
\label{hyper-table}
%\vskip 0.15in
\begin{center}
%\begin{small}
\begin{tabular}{|c|c|c|c|}
\hline
Learning rate & $\beta$ & $\lambda$ & Performance \\ \hline
3e-4          & 1e-1 & 5   & 112.1$\pm$3.4   \\ \hline
3e-4          & 3e-1 & 5   & 109.1$\pm$5.7   \\ \hline
3e-4          & 1e-1 & 2   & 110.9$\pm$6.4   \\ \hline
3e-4          & 3e-1 & 2   & 108.2$\pm$4.2   \\ \hline
1e-4          & 1e-1 & 5   & 111.5$\pm$4.4   \\ \hline
\end{tabular}
%\end{small}
\end{center}
%\vskip -0.1in
\end{table}
\end{document}